\documentclass[lettersize,journal]{IEEEtran}
\usepackage{amsmath,amsfonts}
\usepackage{algorithmic}
\usepackage{array}
\usepackage[caption=false,font=normalsize,labelfont=sf,textfont=sf]{subfig}
\usepackage{textcomp}
\usepackage{stfloats}
\usepackage{url}
\usepackage{verbatim}
\usepackage{graphicx}

\usepackage{color}
\usepackage{hyperref}
\usepackage[capitalise]{cleveref}
\usepackage{amssymb}
\usepackage{algorithm}
\usepackage{booktabs}
\usepackage{makecell}
\usepackage{threeparttable}
\usepackage{multirow}

\newif\ifshowchanges
\showchangesfalse

\newcommand{\change}[1]{\ifshowchanges \textcolor{blue}{#1} \else #1 \fi}
\newcommand{\changetable}{\ifshowchanges \color{blue} \else \fi}
\newcommand{\changefigure}[1]{\ifshowchanges \colorbox{blue}{\fbox{#1}} \else #1 \fi}

\def\BibTeX{{\rm B\kern-.05em{\sc i\kern-.025em b}\kern-.08em
    T\kern-.1667em\lower.7ex\hbox{E}\kern-.125emX}}
\usepackage{balance}

\newcommand{\ie}{{\em i.e.}}

\begin{document}
\title{Exploring Dualistic Meta-Learning to Enhance Domain Generalization in Open Set Scenarios}
\author{Xiran Wang, Jian Zhang, Lei Qi, Yang Gao, Yinghuan Shi
\thanks{
    The Corresponding author is Yinghuan Shi. 
    Xiran Wang, Jian Zhang, Gao Yang and Yinghuan Shi are with the State Key Laboratory for Novel Software Technology, Nanjing University, China. 
    Lei Qi is with the School of Computer Science and Engineering, Southeast University, China. 
    This work was supported by NSFC Project (62536005, 62192783, 62506162), Jiangsu Science and Technology Project (BF2025061, BK20251241), Fundamental and Interdisciplinary Disciplines Breakthrough Plan of the Ministry of Education of China (JYB2025XDXM118), “111 Center” (B26023) and Fundamental Research Funds for the Central Universities (KG202508).
    } \\
    }


\maketitle

\begin{abstract}

Domain generalization learns from multiple source domains to generalize to unseen target domains. However, it often neglects the realistic case of label mismatch between source and target. Open set domain generalization is then proposed to recognize unseen classes in unseen domains.
A simple approach trains one-vs-all classifiers to separate each class and detect outliers as unknown. Yet, the imbalance between few positive samples and many negative samples skews the decision boundary towards the positive ones, leading the model to over-reject out-of-distribution data, even from known classes in unseen domains.
In this paper, we propose a novel meta-learning stategy called dualistic MEta-learning with joint DomaIn-Class matching (MEDIC), which considers implicit gradient matching towards inter-domain and inter-class task splits simultaneously to find optimal boundaries balanced for both domains and classes.
Experimental results show that MEDIC not only outperforms prior methods in open set scenarios, but also maintains competitive close set generalization ability.
Our code is available at   \href{https://github.com/zzwdx/MEDIC-plus}{this link}.
\end{abstract}

\section{Introduction}
\label{sec:intro}
\IEEEPARstart{D}{eep} neural networks have achieved enormous success in a wide range of computer vision tasks, usually assuming that the training and test samples are drawn from the same data distribution and label space.
However, real-world application scenarios often introduce unpredictability, potentially placing the model at risk of performance degradation when the above constraints are not satisfied \cite{li2019research}. 
Domain generalization (DG) \cite{wang2022generalizing} is then motivated as a more realistic setting to deal with data distribution shift, which refers to using multiple source domains 
to obtain a model with the generalization ability that can be directly applied to arbitrary unseen target domains.

Most current domain generalization researches \cite{li2018learning, li2019episodic, zhou2020domain, hemati2023understanding} are based on the assumption of close set recognition, \emph{i.e.}, the classes of the source domains are consistent with those of the target domains. 
However, in practice, the deployed model is often exposed to some new classes that have never been encountered during the training phase \cite{scheirer2012toward}.
For example, 
in medical imaging, some diseases are so rare \cite{griggs2009clinical} that obtaining their training samples is unrealistic.
In close set classification, objects are forced to be assigned into a known class, which introduces potential risks to the model's robustness and security.
To mitigate this issue, it is essential to explore a more practical setting called open set domain generalization (OSDG), which aims to recognize unknown classes while maintaining original classification accuracy of known classes.

\begin{figure}[t]
    \centering
    \includegraphics[width=1\linewidth]{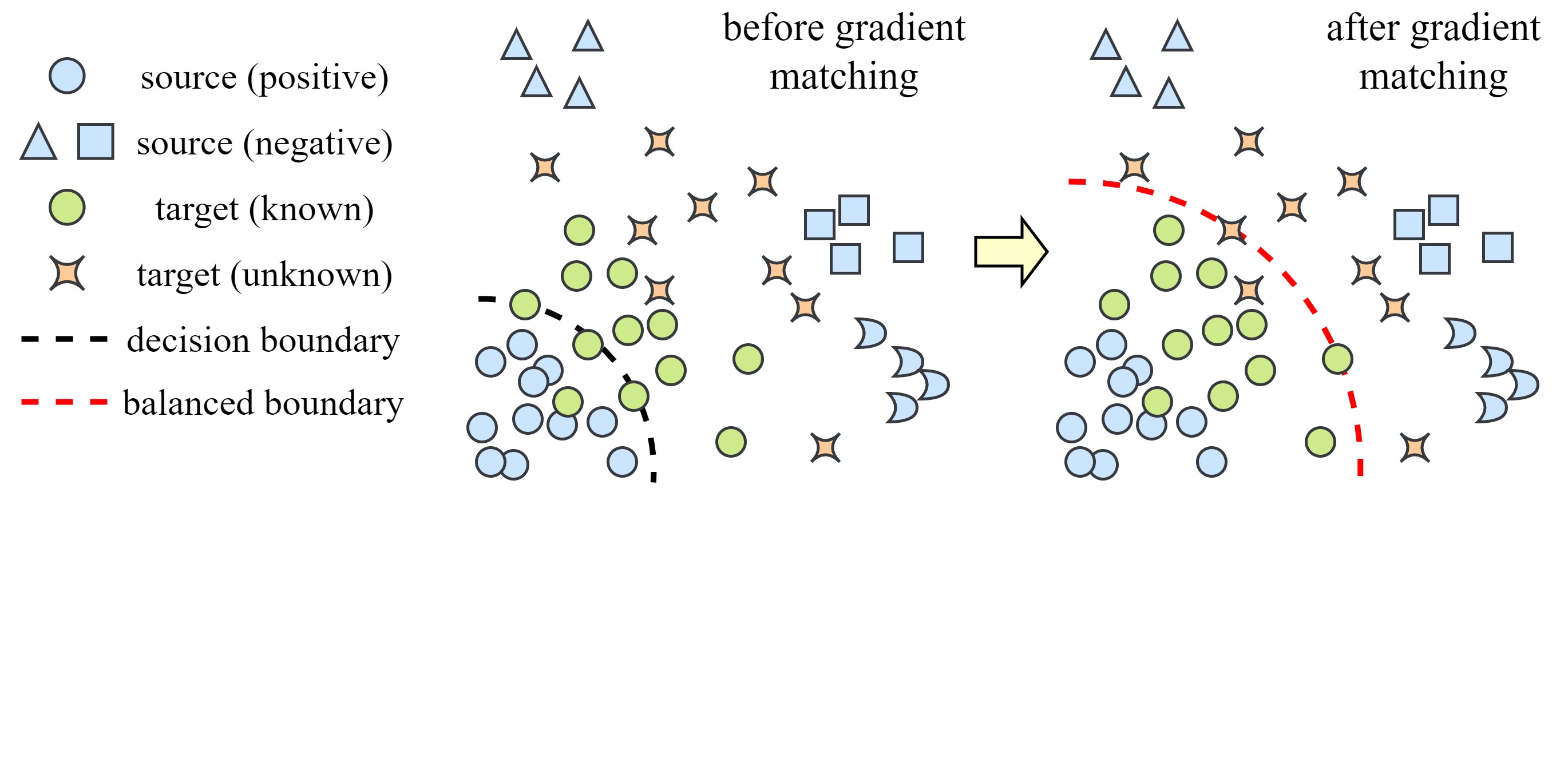}
    \vspace{-0.8in}
    \caption{An example for the variation of decision boundaries of a \emph{one-vs-all} classifier in open set domain generalization.}
    \vspace{-0.1in}
    \label{fig:bias}
\end{figure}

In open set domain generalization \cite{shu2021open, katsumata2021open}, the key is to address both domain shift and category shift simultaneously. However, traditional open set recognition models are not easily applicable for domain generalization tasks because they tend to generate biased decision boundaries. {\emph{i.e.}}, only modeling the training data while neglecting the out-of-distribution samples \cite{guo2021conditional, sun2020conditional}. 
For example, the multi-binary classifier \cite{saito2021ovanet, liu2019separate} consists of multiple one-vs-all binary classifiers to define a decision boundary for each known class. 
If a given sample is classified as negative by all sub-classifiers, it is considered to have a high probability of belonging to unknown classes.
As shown in \cref{fig:bias}, the limited data distribution of the positive samples (\emph{i.e.}, from only one corresponding class) and the more diverse distribution of the negative samples (\emph{i.e.}, from all other classes) can increase the risk of predicting inputs as positive rather than negative. This causes the decision boundary asymmetrically biased to the positive samples, potentially rejecting all out-of-distribution samples as unknown and misclassifying known classes in the unseen target domain.

\begin{figure}[t]
    \centering
    
    \includegraphics[width=0.98\linewidth]{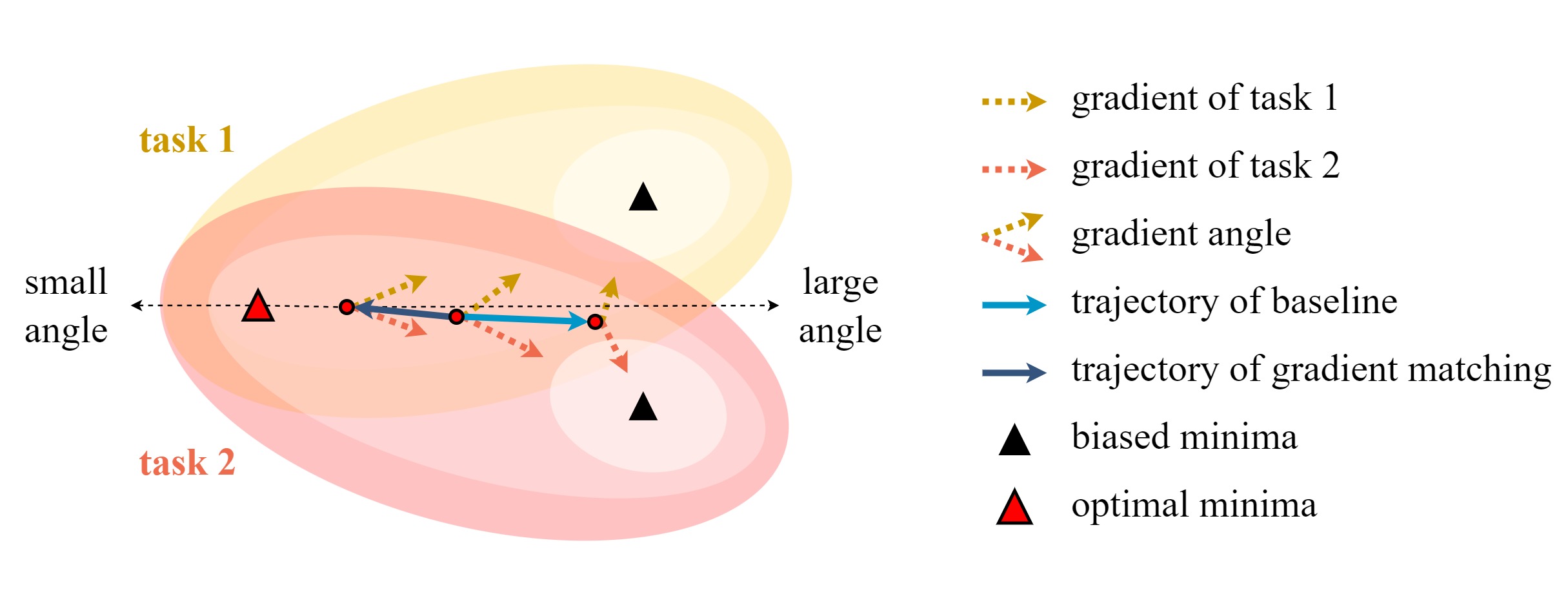}
    \vspace{-0.1in}
    \caption{Previous research \cite{shi2021gradient} has demonstrated that the large angle between gradients of two tasks introduces contradictions in optimization.}
    \vspace{-0.1in}
    \label{fig:gradient}
\end{figure}

To establish a balanced decision boundary across domains and classes, our attention turns to meta-learning \cite{hospedales2021meta}, a simple yet effective approach for handling domain shift. Prior work on meta-learning-based domain generalization \cite{li2018learning, shi2021gradient} seeks an optimal balance among domains by matching gradients across tasks sampled from them.
This domain-wise meta-learning can mitigate the risk of exhibiting excessive bias towards particular domains.
As shown in \cref{fig:gradient}, the rationale is that if the angles between the gradients are small, which implies that optimizing one task does not interfere with other tasks, then it is possible to achieve a win-win outcome by optimizing their combined gradient. 
In contrast, large angles between gradients indicate conflicting objectives, where updating one task can adversely impact the optimization procedure of others.

We propose learning positive and negative samples in such a balanced way, to place the decision boundary at the middle zone of them, attaining a more rational separation between known and unknown classes in the target domain.
Concretely, we introduce a novel meta-learning strategy called \emph{dualistic MEta-learning with joint DomaIn-Class matching (MEDIC)}.
Instead of simply adding extra iterations for inter-domain or inter-class meta-learning, we take a step further to achieve gradient matching between domains and classes simultaneously.
For tasks selected from different domains, we additionally split and recombine them at the category level to construct inter-class pairs.
By matching the gradients of these recombined tasks, we expect the model to not only generalize well across domains, but also to grasp a more precise understanding of class-wise relationships, which is beneficial for both close set generalization and open set recognition.
This article extends our original work \cite{wang2023generalizable} from an initial insight to a generalized framework with accompanying theory and experiments.

\begin{itemize}
\item{We investigate inter-class gradient matching for open set domain generalization. The method is introduced from a special case (\emph{i.e.}, two steps per inner loop) to a general one (\emph{i.e.}, multiple steps per inner loop) with an integrated task scheduling strategy for hard class pairs.}

\item{We provides a more precise theoretical analysis of step-wise gradient matching compared to original proofs \cite{nichol2018first} \cite{shi2021gradient},
as eliminating reliance on mathematical expectation.
Our strategy could achieve task-wise gradient matching close to its maximum value with fewer steps.
}
\item{Extensive experiments show that our method not only outperforms several state-of-the-art methods in the open set scenario, but also maintains remarkable accuracy in the conventional domain generalization setting.}
\end{itemize}

\section{Related Work}
\label{sec:related}

\subsection{Domain Generalization}
\label{subsec:related-dg}

Domain Generalization (DG) is intended to train a model on multiple source domains that can generalize to unseen target domains without extra retraining process. Existing methods mainly focus on three directions: (i) \textbf{Feature Representation}, learning domain-invariant features through techniques such as domain adversarial learning \cite{li2018domain, ganin2016domain, sicilia2023domain, chen2024domain}, invariant risk minimization \cite{arjovsky2020invariant, ahuja2021invariance}, or causality-based feature disentangling \cite{chattopadhyay2020learning, lv2022causality}. (ii) \textbf{Data Augmentation}, enhancing training diversity via domain transfer, mixing or Fourier transforms \cite{zhou2020domain, wang2020heterogeneous, xu2021fourier, guo2023aloft}, adversarial generation \cite{li2021progressive, zhou2020learning}, or stochastic noise injection \cite{li2021simple, wang2022feature}. (iii) \textbf{Learning Strategy}, applying meta-learning \cite{zhang2022mvdg, dou2019domain, balaji2018metareg}, ensemble learning \cite{zhou2021domain, cha2021swad, arpit2022ensemble, tian2023privacy}, or regularization \cite{huang2020self, wang2022domain, shi2021gradient, mansilla2021domain} where some can be efficiently achieved via meta-learning.

\begin{table}[t]
\caption{Comparison of target domains under different settings.}
\vspace{-0.05in}
\centering
\begin{threeparttable}
\resizebox{1.\linewidth}{!}{
\begin{tabular}{lccc}
\toprule
\textbf{\makecell[l]{Problem Setting}} & 
{\makecell[c]{Distribution \\ of Data}} & 
{\makecell[c]{Label \\ Space}} & 
{\makecell[c]{Participation \\ in Training}} \\
\midrule
Domain Adaptation \cite{wang2018deep} & $\mathcal{Q}$ & $\mathcal{C}$ & \checkmark \\
Domain Generalization \cite{wang2022generalizing} & $\mathcal{Q}$ & $\mathcal{C}$ & $\times$ \\
Open Set Recognition \cite{geng2020recent} & $\mathcal{P}$ & $\mathcal{C \cup U}$ & $\times$ \\
Open Set Domain Generalization \cite{shu2021open} & $\mathcal{Q}$ & $\mathcal{C\cup U}$ & $\times$ \\
\bottomrule
\end{tabular}
}
\begin{tablenotes}
	\item[1] $\mathcal{P}$ and $\mathcal{C}$ are data distribution and label space of source domains. 
    \item[2] $\mathcal{Q}$ is the unseen data distribution and $\mathcal{C}\cap \mathcal{U}=\varnothing$.
\end{tablenotes}
\end{threeparttable}
\vspace{-0.1in} 
\label{tab: pbs-cmp}
\end{table}
\subsection{Open Set Recognition}
\label{subsec:related-osr}

Open Set Recognition (OSR) focuses on detecting novel classes not included in the training set. Based on the use of additional data, existing methods can be classified into two categories.
(i) \textbf{Artificial Classes.} Some methods \cite{dhamija2018reducing, hendrycks2018deep} augment the training data with auxiliary classes to improve distinction among known classes, but their effectiveness depends heavily on the quality of these samples. Others \cite{ge2017generative, neal2018open} propose to use generative models to guess unknown class samples, yet the resulting images are often low in quality and far from realistic, making them less effective on complex datasets \cite{kong2021opengan}.
(ii) \textbf{Discriminative Models.} OpenMax \cite{bendale2016towards} replaces the softmax layer and estimates unknown probabilities using EVT \cite{smith1990extreme}. Self-supervised methods \cite{oza2019c2ae, yoshihashi2019classification, yue2022spectral} leverage reconstruction errors, as they believe that known-class samples are usually reconstructed more accurately than unknown ones. Metric learning \cite{chen2021adversarial, guo2021conditional, lu2022pmal} is also widely used to enhance feature discrimination. However, these approaches often misclassify all out-of-distribution samples as unknown, limiting their direct application to domain generalization.
\subsection{Meta-Learning}
\label{subsec:related-ml}

\change{
Meta-learning, also known as learning to learn \cite{thrun2012learning, al2021data}, aims to equip models with the ability to generalize across tasks by finding an initialization that can be quickly adapted with minimal updates. The model-agnostic meta-learning (MAML) \cite{finn2017model} and first-order meta-learning (Reptile) \cite{nichol2018first} divide the model learning process into inner and outer loops.
The inner loop is for task-specific adaptation, while the outer loop seeks a globally optimal initialization for the task in the inner loop.
In domain generalization, meta-learning has been applied to balance optimization across diverse domains \cite{li2018learning, balaji2018metareg, shi2021gradient}. MLDG \cite{li2018learning} simulats domain shifts via meta-train and meta-test splits. Fish \cite{shi2021gradient} introducs a first-order strategy to reduce computational cost. Unlike these domain-level strategies, our approach further samples tasks at the class level to prevent biased decision boundaries and better distinguish known from unknown classes in the target domain.}
\subsection{Open Set Domain Generalization}
\label{subsec:related-osdg}

\change{
Open Set Domain Generalization (OSDG), summarized in \cref{tab: pbs-cmp}, aims to address both domain and class shifts. Previous studies mainly focus on training highly discriminative models \cite{shu2021open, katsumata2021open, yang2022one} or rejecting unknown classes at test time \cite{chen2023activate}.
A key limitation of prior methods is their separate treatment of the two shifts.
For example, DAML \cite{shu2021open}, based on domain augmentation and meta-learning, primarily targets the data shift between source domains.
CrossMatch \cite{zhu2021crossmatch} adopts consistency regularization between the close set classifier and multi-binary classifier without considering the domain shift. And our object is to tackle both shifts within a unified framework.}

\change{
More recent methods \cite{singha2024unknown,bose2025meta,gupta2025osloprompt} use stronger models to improve open set performance. Some employ stable diffusion \cite{rombach2022high} to synthesize unknown classes, while some generate class descriptions using rules or GPT-4o \cite{achiam2023gpt} and integrate them into CLIP \cite{radford2021learning}.
Concurrently, several meta-learning methods have been developed based on MEDIC \cite{wang2023generalizable}. 
L2OT \cite{li2025learning} adds a regularization term to penalize similar distributions between different classes.
EBiL-HaDS \cite{peng2024advancing} incorporates noisy samples into training and proposes a scheduler to select challenging domain–class pairs to separate tasks. 
HyProMeta \cite{peng2026mitigating} detects noisy samples using class prototypes and split meta-train and test sets according to clean and noisy samples. These studies show the adaptability of MEDIC, and the MEDIC++ proposed in this paper is a more generalized baseline, of which MEDIC can be regarded as a special case.}

\section{Method}
\label{sec:method}
\cref{subsec:method-preliminary} discuss some definitions in open set domain generalization and the training process of meta-learning. 
\cref{subsec:method-medic} and \cref{subsec:method-medic++} present our MEDIC++ framework.
\cref{subsec:method-sample} establishes the adaptive task sampling strategy.
Finally, \cref{subsec:method-loss} and \cref{subsec:method-inference} introduce the multi-binary classifier and its inference methodology.
\begin{figure*}[tp]  
    \centering
    \includegraphics[width=0.96\linewidth]{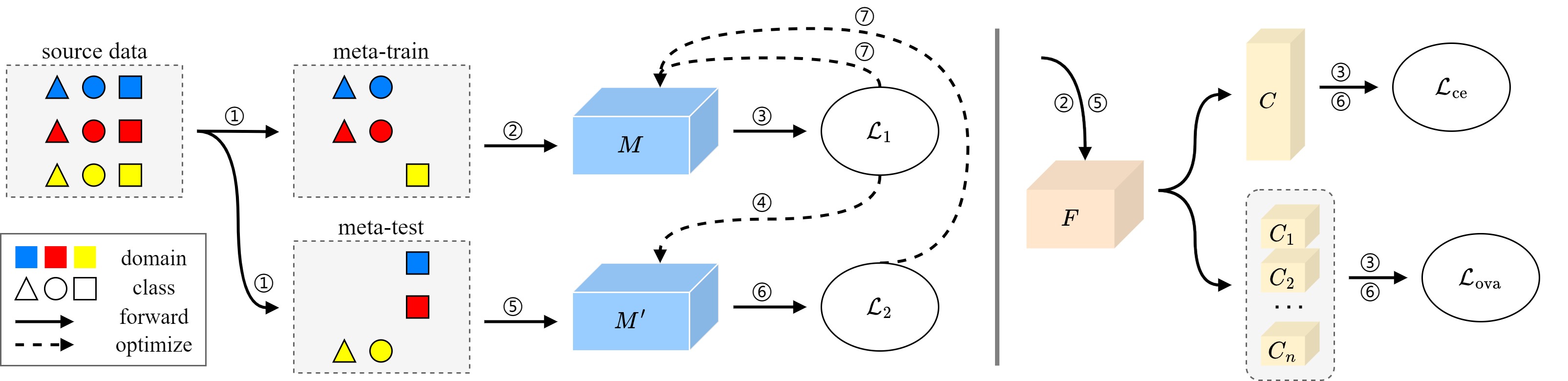}
    \caption{
    Overview of MEDIC during one training iteration. 
    $M$ is the overall model, and the right figure represents its internal structure.
    The numbers denote the sequence of data flow (solid arrows) and model updates (dashed arrows) respectively. 
    } 
    \vspace{-0.1in}
    \label{fig:medic}
\end{figure*}
\subsection{Preliminary}
\label{subsec:method-preliminary}

\noindent \textbf{Problem definition.} For open set domain generalization, we are provided with $S$ source domains $\mathcal{S} = \lbrace \mathcal{D}_1,\mathcal{D}_2,...,\mathcal{D}_S\rbrace$ with a label space $\mathcal{C}$ and $T$ unseen target domains $\mathcal{T} = \lbrace \mathcal{D}_{S+1},\mathcal{D}_{S+2},...,\mathcal{D}_{S+T}\rbrace$ with an extended label space $\mathcal{C}\cup\mathcal{U}$ which satisfies $\mathcal{C}\cap \mathcal{U}=\varnothing$. 
The $s$-th domain consisting of $N_s$ samples is represented as $\mathcal{D}_s=\lbrace(x^s_i,y^s_i)\rbrace_{i=1}^{N_s}$, where $x^s_i$ denotes the $i$-th sample and $y^s_i$ can take values from $\mathcal{C}$ for $\mathcal{S}$ and  $\mathcal{C}\cup\mathcal{U}$ for $\mathcal{T}$, which refers to the corresponding label in the source or target domains.
Our objective is to utilize these source domains $\mathcal{S}$ to develop a model that can generalize to any unseen domain $\mathcal{T}$.
Below are several key terms.

\vspace{0.3em}

\noindent \change{\textbf{Task splits}, also termed task partition of a dataset $\mathcal{S}$ is defined as a collection of non-empty subsets 
$\{\mathcal{S}_1,\mathcal{S}_2,\dots,\mathcal{S}_t\}$ such that
$\mathcal{S}_i \cap \mathcal{S}_j = \varnothing \ (i \neq j)$, 
and $\bigcup_{i=1}^{t} \mathcal{S}_i = \mathcal{S}$. 
That is, the subsets are mutually disjoint and their union constitutes the entire dataset $\mathcal{S}$.
For example, consider a dataset with domain labels $\{a,b\}$ and class labels $\{1,2\}$:
$
\mathcal{S} = \{(a,1), (a,2), (b,1), (b,2)\}.
$
Two valid partitions can be constructed in different ways. 
Partitioning by domain yields
$
\big\{\{(a,1),(a,2)\},\{(b,1),(b,2)\}\big\},
$
while partitioning by class yields
$
\big\{\{(a,1),(b,1)\},\{(a,2),(b,2)\}\big\}.
$
These represent two different criteria for dividing the same dataset.
In this paper, a task refers to a mini-batch of samples drawn from a specific subset of such a partition.}

\vspace{0.3em}

\noindent \change{\textbf{MLDG-like meta-learning} \cite{li2018learning} requires to split the source domains $\mathcal{S}$ into the meta-train set $\mathcal{S}_{\mathcal{F}}$ and meta-test set $\mathcal{S}_{\mathcal{G}}$, ensuring that  $\mathcal{S}_{\mathcal{F}} \cup \mathcal{S}_{\mathcal{G}}=\mathcal{S}$ and $\mathcal{S}_{\mathcal{F}} \cap \mathcal{S}_{\mathcal{G}}=\varnothing$. 
The losses on these two sub-datasets $\mathcal{S}_{\mathcal{F}}$ and $\mathcal{S}_{\mathcal{G}}$ are defined as $\mathcal{F}(\Theta)$ and $\mathcal{G}(\Theta)$ respectively, with $\Theta$ representing the parameters of the training model.
The optimization is conducted in a bi-level manner.
First, the parameters $\Theta$ are updated to $\Theta'$ with the loss of $\mathcal{F}(\Theta)$.
Then the loss of $\mathcal{G}(\Theta')$ is combined with  $\mathcal{F}(\Theta)$ to update the model's original parameters $\Theta$.}

\vspace{0.3em}

\noindent \change{\textbf{Reptile-like meta-learning} \cite{nichol2018first} segments the optimization process into an inner loop and an outer loop.
Each iteration can be summarized as follows: A task consists of a batch of data sampled from a particular data distribution, and a step aggregates several tasks to create a larger batch.
During the inner loop, the model is sequentially updated with steps to reach parameters $\hat{\Theta}$.
Then in the outer loop, the original parameters $\Theta$ are updated in the direction of $\hat{\Theta}-\Theta$.
The optimization of MLDG-like meta-learning could also be approximated using the two-step reptile scheme, with one step for meta-train and the other for meta-test. In this case, $\Theta-\hat{\Theta} = \alpha\mathcal{F}(\Theta)+\beta\mathcal{G}(\Theta')$ where $\alpha$ and $\beta$ are inner loop learning rates.}

\vspace{0.3em}

\noindent \change{\textbf{Gradient matching} refers to searching parameters where the gradient directions induced by different tasks are well aligned. 
The goal is to ensure that optimizing one task does not cause significant interference with others, thereby promoting consistent improvement.
Directly enforcing gradient matching as an explicit regularization term would require computing second-order derivatives, which introduces substantial computational overhead. 
In practice, meta-learning is commonly adopted to achieve implicit gradient matching. 
A detailed discussion will be provided in the following sections.
}

\begin{figure*}[t]
    \centering
    \includegraphics[width=0.98\linewidth]{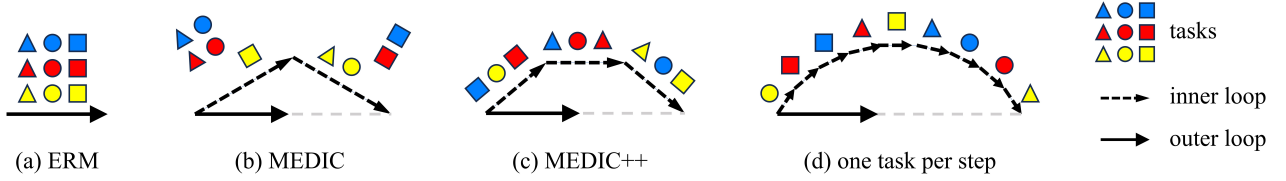}
    \vspace{-0.1in}
    \caption{Comparison of different learning strategies. (a) A single step. (b) (c) Multiple tasks per step. (d) Maximum steps.
    A greater number of steps implies a larger number of task-wise gradient-matching.
    Please turn to \cref{fig:norm}(d) for more details.} 
    \vspace{-0.1in}
    \label{fig:medic++}
\end{figure*}

\subsection{MEDIC as a Special Case}
\label{subsec:method-medic}

To clarify the mechanism of gradient matching across multiple attributes (\ie, domain and class), we begin by introducing a special case
called MEDIC, where both of the domains and classes are separated into two parts, yielding four tasks in total, and each step selects two of them.

Given two sub-datasets $\mathcal{S}_{\mathcal{F}}$ and $\mathcal{S}_{\mathcal{G}}$, along with their corresponding loss functions $\mathcal{F}(\Theta)$ and $\mathcal{G}(\Theta)$, our objective is to reach a consensus on their gradients ${\mathcal{F}'(\Theta)}$ and ${\mathcal{G}'(\Theta)}$ to ensure an unbiased optimization direction for both of them. 
The underlying principle is that if the angle between the directions of ${\mathcal{F}'(\Theta)}$ and ${\mathcal{G}'(\Theta)}$ is small which means optimizing one task does not adversely affect the other, then updating with their combined gradient (\emph{i.e.}, the sum of gradients in practice) can yield enhanced performance.
Conversely, if the angle between their gradients is large which indicates a conflict between these two tasks, then updating one task would lead to an inferior optimization process for the other.
The core idea of gradient matching is to find an area in the parameter space where the angle between the gradients of $\mathcal{S}_{\mathcal{F}}$ and $\mathcal{S}_{\mathcal{G}}$ is minimized, which can be accomplished by maximizing the dot product of ${\mathcal{F}'(\Theta)}$ and ${\mathcal{G}'(\Theta)}$.
Moving the model towards this region, $\mathcal{S}_{\mathcal{F}}$ and $\mathcal{S}_{\mathcal{G}}$ will converge on gradient direction, where both tasks benefit from shared updates rather than competing each other.

Current gradient-based domain generalization methods typically treat $\mathcal{S}_{\mathcal{F}}$ and $\mathcal{S}_{\mathcal{G}}$ as separate domains in order to find an optimization direction only among domains \cite{li2018learning, shi2021gradient}.
However, these methods often overlook the inter-class relationships, which are important in open set scenarios. Instead of simply adding extra iterations to mitigate biased prediction between classes, we propose a novel meta-learning strategy that performs gradient matching across both inter-domain and inter-class splits simultaneously, which aims to learn generalizable decision boundaries that maintain balance across all tasks.

As illustrated in \cref{fig:medic}, for tasks $\mathcal{S}_{\mathcal{F}}$ and $\mathcal{S}_{\mathcal{G}}$ sampled from different domains, we further divide them into $\mathcal{S}_{\mathcal{F}_1}, \mathcal{S}_{\mathcal{F}_2}$ and $\mathcal{S}_{\mathcal{G}_1}, \mathcal{S}_{\mathcal{G}_2}$ by class and define their loss functions as $\mathcal{F}_1, \mathcal{F}_2$ and $\mathcal{G}_1, \mathcal{G}_2$. 
The label spaces of $\mathcal{S}_{\mathcal{F}_1}$ and $\mathcal{S}_{\mathcal{F}_2}$, as well as $\mathcal{S}_{\mathcal{G}_1}$ and $\mathcal{S}_{\mathcal{G}_2}$ are both disjoint.
Besides, we require $\mathcal{S}_{\mathcal{F}_1}$ and $\mathcal{S}_{\mathcal{G}_1}$ to share the same label space, and likewise for $\mathcal{S}_{\mathcal{F}_2}$ and $\mathcal{S}_{\mathcal{G}_2}$. 
To simultaneously apply gradient matching between domains and classes, we utilize $(\mathcal{S}_{\mathcal{F}_1}, \mathcal{S}_{\mathcal{G}_2})$ as meta-train set and $(\mathcal{S}_{\mathcal{F}_2}, \mathcal{S}_{\mathcal{G}_1})$ as meta-test set.  
The final meta-objective function of MEDIC is defined as follows:
\begin{equation}
\label{eq:medic-objective}
\mathop{\rm argmin}_{\Theta}\, [{\mathcal{F}_1(\Theta)} + {\mathcal{G}_2(\Theta)}] + \beta
[{\mathcal{F}_2(\hat{\Theta})} + {\mathcal{G}_1(\hat{\Theta})}].
\end{equation}
This objective approximates the two-step reptile scheme, with one step for the meta-train set $(\mathcal{S}_{\mathcal{F}_1}, \mathcal{S}_{\mathcal{G}_2})$ and the other for the meta-test set $(\mathcal{S}_{\mathcal{F}_2}, \mathcal{S}_{\mathcal{G}_1})$, 
where $\beta$ controls the weight between the two meta sets and $\hat{\Theta}$ is the optimized model parameters on the meta-train set with learning rate $\alpha$: 
\begin{equation}
{\hat{\Theta}} = \Theta - \alpha({\mathcal{F}_1'(\Theta)} + {\mathcal{G}_2'(\Theta)}).
\end{equation}
To validate MEDIC's capability to perform gradient matching between domains and classes at the same time, similar to the analysis in \cite{li2018learning}, we conduct a first order Taylor expansion for the second term in \cref{eq:medic-objective}:
\begin{align}
    {\mathcal{F}_2(\hat{\Theta})} &= {\mathcal{F}_2(\Theta)} - \alpha (
{\mathcal{F}_1'(\Theta)} + {\mathcal{G}_2'(\Theta)}) \cdot {\mathcal{F}_2'(\Theta)}, \\
    {\mathcal{G}_1(\hat{\Theta})} &= {\mathcal{G}_1(\Theta)} - \alpha (
{\mathcal{F}_1'(\Theta)} + {\mathcal{G}_2'(\Theta)}) \cdot {\mathcal{G}_1'(\Theta)},
\end{align}
and the objective function becomes:
\begin{equation}
\label{eq:taylor-objective}
\begin{split}
& \mathop{\rm argmin}_{\Theta}\, [{\mathcal{F}_1(\Theta)} + {\mathcal{G}_2(\Theta)} + \beta
({\mathcal{F}_2(\Theta)} +  {\mathcal{G}_1(\Theta)})] \\ &- \beta \alpha [({\mathcal{F}_1'(\Theta)} + {\mathcal{G}_2'(\Theta)})\cdot(
{\mathcal{F}_2'(\Theta)} +
{\mathcal{G}_1'(\Theta)})].
\end{split}
\end{equation}
The first term of \cref{eq:taylor-objective} involves optimizing the model with the expected losses of each task, while the second term is the product of gradient sums. By expanding this part we derive the following regularization term:
\begin{equation}
\label{eq:regularization}
\mathcal{L}_{\rm reg} = -(\mathcal{F}_1' \cdot \mathcal{F}_2' + 
\mathcal{F}_1' \cdot
\mathcal{G}_1' +
\mathcal{G}_2' \cdot
\mathcal{F}_2' +
\mathcal{G}_2' \cdot
\mathcal{G}_1').
\end{equation}
Note that we omit parameters $\Theta$ for the sake of simplicity.
As previously discussed, maximizing the dot product of gradients can regularize the optimization process to match the updating directions of different tasks.
Taking task $\mathcal{S}_{\mathcal{F}_1}$ as an example, the multiplier $\mathcal{S}_{\mathcal{G}_1}$ contains the same classes but from different domains, whereas $\mathcal{S}_{\mathcal{F}_2}$ includes different classes from the same domain, the two factors in any of the gradient products are either from different domains or different classes to enable domain-wise and class-wise matching simultaneously.
In contrast to conventional methods primarily concerned with inter-domain relationships, the dot product between $\mathcal{S}_{\mathcal{F}_1}$ and $\mathcal{S}_{\mathcal{F}_2}$ bridges the gap in class-wise gradient matching inside each domain, which enables fine-grained model optimization to learn more rational decision boundaries.

\begin{figure*}[tp]  
    \centering   
    \includegraphics[width=1\linewidth]{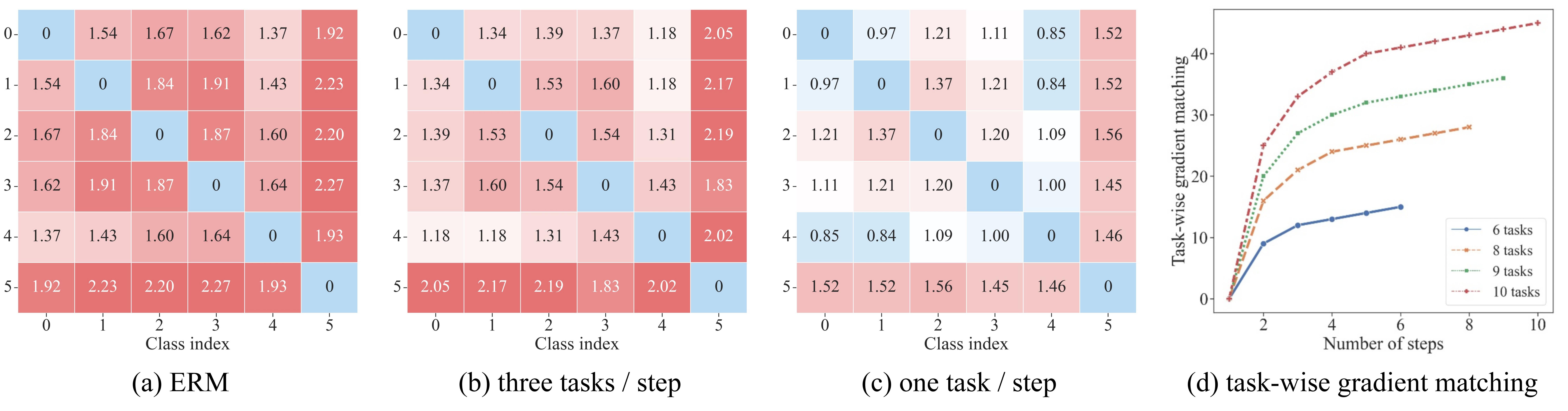}
    \vspace{-0.25in}
    \caption{
    (a) (b) (c) Standardized mean differences of class feature pairs. Warmer colors indicate more distinct features.
    It can be observed that the one task per step strategy leads to a convergence of their feature representations, consequently disrupting the classification performance of the model.
    (d) As the number of steps increases, the rate of increase in task-wise gradient matching gradually decreases, eventually converging to 0.
     }
    \vspace{-0.1in}
    \label{fig:norm}
\end{figure*}

\subsection{The MEDIC++ Framework}
\label{subsec:method-medic++}
MEIDC++ is a flexible meta-learning strategy that involves several tasks at each step. 
It can be extended into a general learning framework by creating additional tasks with finer division between domains and classes, while the number of tasks per step can also be customized as we need.

Existing meta-learning-based DG methods usually overlook the distinction between tasks and steps, with a preference for sampling only one task at each step. 
Fish \cite{shi2021gradient} theoretically demonstrates the effectiveness of this approach in pairing tasks with one another, allowing for maximum gradient matching across all tasks.
However, this strategy can become complicated when dealing with a large quantity of tasks, as the number of steps increases linearly with them.
Besides, the batch normalization-based models over training only normalize the mean and variance of each single batch. As shown in \cref{fig:norm}, updating with only one task per step leads to similar statistics across classes, making their features indistinguishable and may negatively affect classification performance.
Conversely, if we just take a single large step that encompasses all the tasks, the number of paired steps will be reduced to zero, leading to no gradient matching between any tasks.

We argue that taking multiple steps is essential for task-wise gradient matching, but there is no need for a large number of them.
To this end, we propose our MEDIC++ as follows:
During the inner loop, the dataset is partitioned into $t$ tasks $\lbrace \mathcal{S}_{\mathcal{H}_1}, \mathcal{S}_{\mathcal{H}_2}, ..., \mathcal{S}_{\mathcal{H}_t}\rbrace$ at both the domain and the class levels, with their loss functions represented as $\lbrace\mathcal{H}_1, \mathcal{H}_2, ..., \mathcal{H}_t\rbrace$.
We define the number of steps as $n$ and the corresponding loss functions as $\lbrace\mathcal{L}_1, \mathcal{L}_2, ..., \mathcal{L}_n\rbrace$.
The number of tasks included in the $i$-th step is denoted as $m_i$ that satisfies $\sum_{i=1}^{n}m_i = t$.
In each step, tasks are randomly sampled without repetition for gradient update, with the condition that $n$ is less than $t$.
We assume a uniform distribution of tasks per step, \emph{i.e.}, 
tasks are distributed as $\lfloor \frac{t}{n} \rfloor$ in some steps while $\lceil \frac{t}{n} \rceil $ in others.
The loss function of the $i$-th step is expressed as: 
\begin{equation}
\label{eq:step-loss}
\mathcal{L}_i = \sum_{k=1}^{m_i}\mathcal{H}_{a_i^k}, 
\end{equation}
where $a_i^k$ is the index of the $k$-th task chosen in the $i$-th step.
Following the inner loop, the model's parameters are updated from $\Theta$ to $\hat{\Theta}$, and we adopt the final parameters as
$\Theta \leftarrow \Theta + \epsilon(\hat{\Theta}-\Theta)$, 
where $\epsilon$ can be considered as the negative learning rate of the outer loop.
This reptile-like meta-learning \cite{nichol2018first, shi2021gradient} conducts pairwise gradient matching between each step with scaling factor $\gamma$: 
\begin{equation}
\label{eq:reptile-objective}
\mathop{\rm argmin}_{\Theta}\, \sum_{i=1}^{n}\mathcal{L}_i(\Theta)-\gamma\sum_{i, j \in \mathcal{N}}^{i\neq j}\mathcal{L}_i'(\Theta) \cdot \mathcal{L}_j'(\Theta),
\end{equation}
where $\mathcal{N}$ is the indices of $\lbrace 1,2,...,n\rbrace$, thus the number of step-wise gradient matching can be calculated as $\frac{n(n-1)}{2}$.
We provide a more precise proof of \cref{eq:reptile-objective} and please refer to \cref{subsec:appendix-a} for more details.
When substituting \cref{eq:step-loss} and its derivative into \cref{eq:reptile-objective}, we can obtain the objective:
\begin{equation}
\label{eq:task-objective}
\mathop{\rm argmin}_{\Theta}\, \sum_{i=1}^{t}\mathcal{H}_i(\Theta)-\gamma\sum_{i, j \in \mathcal{N}}^{i\neq j} \left\{ \sum_{k=1}^{m_i}\mathcal{H}_{a_i^k}'(\Theta) \cdot \sum_{k=1}^{m_j}\mathcal{H}_{a_j^k}'(\Theta) \right\},
\end{equation}
implying that matching gradients between steps is equivalent to matching gradients across all inter-step tasks.

\change{
Considering the number of task-wise gradient matching as $t_n$, we show in \cref{subsec:appendix-c} that $t_n$ is positively correlated with the step count $n$. 
By using $\frac{t}{n}$ to approximate the number of task-wise gradient matching per step, we could derive an estimate for $t_n$ as:
\begin{equation}
\label{eq:--}
f(n) = \frac{n(n-1)}{2} \cdot (\frac{t}{n})^2 = \frac{1}{2}t^2 \cdot (1-\frac{1}{n}).
\end{equation} 
Note that if $t$ is divisible by $n$, then $t_n$ and $f(n)$ coincide. The derivative of $f(n)$ is:
\begin{equation}
\label{eq:---}
f'(n) = \frac{t^2}{2n^2},
\end{equation}
which indicates that when $n$ is small, there is a rapid increase in task-wise gradient matching.
However, as $n$ becomes larger, the growth rate converges to zero.
As shown in \cref{fig:norm}(d), if we consider a total of $9$ tasks, $t_n$ reaches its maximum value of $36$ when $n$ equals $9$, while even with $n$ as small as $3$, $t_n$ still achieves a substantial value of $27$, revealing that the inner loop doesn't need to be excessively long. 
In practice, a relatively small number of steps is sufficient to obtain gradient matching $t_n$ close to its maximum value.
Furthermore, since we perform additional task sampling at the class level, updating gradients with multiple tasks allows unified batch normalization across diverse classes, which enables the model to capture specific spatial distribution of each class in the feature space. }

\vspace{0.3em}

\noindent \textbf{Remark 1.} We believe that the rationality behind MEDIC++ can draw inspiration from the sampling strategies of modern gradient descent algorithms, as the sequence of steps mirrors such gradient update operation during each inner loop.
Please be aware that this is merely an informal analogy and should be taken much less seriously than the previous analysis.

We begin by comparing the three task sampling strategies in \cref{fig:medic++} with gradient descent algorithms \cite{ruder2016overview}, assuming that the intra-task samples can be interpreted as a collective data point due to their greater similarity in gradient patterns compared to the inter-task samples. 
In this case, the single-step strategy is similar to batch gradient descent (BGD), which utilizes all data points for each gradient update.
The single-task-per-step strategy resembles a narrower variant of stochastic gradient descent (SGD), involving a single data point per gradient step.
Our MEDIC++, following a multiple-task-per-step strategy, is analogous to mini-batch gradient descent (MBGD) with each gradient computed using a small batch of randomly selected data points.
Modern gradient descent algorithms often prefer MBGD, which aims to seek a balance between exploration and stability \cite{keskar2016large, hinton2012neural, jastrzkebski2017three}, so opting for an multiple-task-per-step approach seems to be a wise choice.

\vspace{0.3em}

\noindent \change{\textbf{Remark 2.} We recommend that one essential contribution of MEDIC++ is to broaden the notion of a task. In domain generalization, it is natural to treat each domain as a separate task, since source domains are predefined. However, if we instead regard the entire training set as a single global distribution, it can be partitioned into tasks along multiple dimensions. We explore class-wise partitioning as an alternative and provide evidence of its effectiveness. In this sense, task can be a design choice rather than an intrinsic property of the dataset, which needs not be limited to explicit domain or class splits.
More broadly, our work suggests that any dataset capable of being decomposed into multiple sub-distributions may benefit from the learning paradigm of MEDIC++.}

\subsection{Adaptive Task Sampling}
\label{subsec:method-sample}

\begin{figure}[t]
    \centering
    \includegraphics[width=0.85\linewidth]{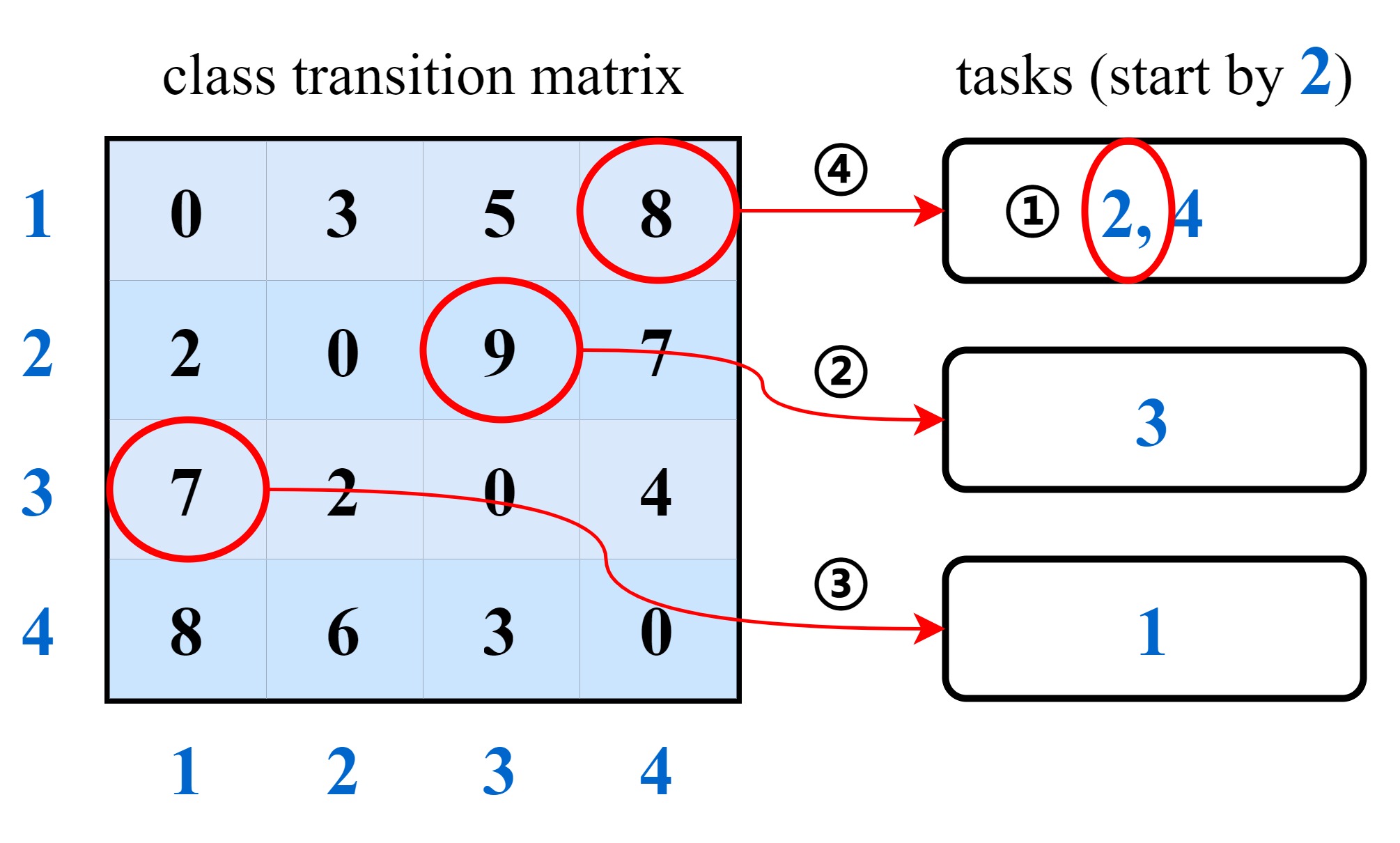}
    \vspace{-0.1in}
    \caption{An example of adaptive task sampling. Initially, class 2 is randomly selected and assigned to task 1. Then the next class 3 is selected based on the transition probabilities without replacement and assigned to task 2.}
    \vspace{-0.1in}
    \label{fig:transition}
\end{figure}

We introduce an adaptive task sampling strategy to enhance the frequency of gradient matching between easily confusable classes. Inspired by the method in \cite{liu2020adaptive}, we first construct an asymmetric class transition probability matrix for each domain, where $p_{ij}$ represents the average prediction probability of misclassifying class $i$ as class $j$:
\begin{equation}
\label{eq:-1}
p_{ij}=\frac{p(j|c=i)}{\sum_{k\neq i}p(k|c=i)}.
\end{equation}
As shown in \cref{fig:transition}, we start by randomly selecting a class. Then, based on the transition probabilities from the selected class to others, we sample the next class without replacement. This process is repeated, with the sampled classes cyclically assigned to different tasks. The rationale behind is that a higher transition probability from class $i$ to class $j$ suggests that the decision boundary between these classes is more likely to be biased, thereby increasing the frequency of gradient matching between them. Since gradient matching only occurs across different tasks, where the adjacent classes are always placed, it is reasonable to use the class transition probabilities to guide the sampling of the next class.

\subsection{Open Set Loss Function}
\label{subsec:method-loss}

For open set recognition, we adopt a multi-binary classifier \cite{saito2021ovanet} to serve as a supplement to the close set classifier.
As illustrated in \cref{fig:medic}, the proposed classifier consists of $|\mathcal{C}|$ one-vs-all classifiers, with each classifier trained to detect whether a given sample belongs to its corresponding class.
Let $p(\hat{y}^k|x)$ denote the output probability that an instance $x$ is an inlier of the $k$-th sub-classifier.
For a given sample $(x, y)$, its loss on the multi-binary classifier can be formulated as:
\begin{equation}
\label{eq:ovaloss}
\mathcal{L}_{\rm ova}(x, y) = -{\rm log}(p(\hat{y}^y|x)) - \mathop{\rm min}_{j \neq y}\,{\rm log}(1-p(\hat{y}^j|x)).
\end{equation}
The second term denotes that it updates only the most challenging binary classifier when used as a negative sample.
We simply adopt this loss and train a close set classifier using cross-entropy loss denoted as $\mathcal{L}_{\rm ce}$.
Then the overall open set loss function can be defined as follows:
\begin{equation}
\label{eq:overalloss}
\mathcal{L}_{\rm all} = \mathcal{L}_{\rm ce} + \mathcal{L}_{\rm ova}.
\end{equation}
\cref{eq:overalloss} is employed as the objective for each task in our meta-learning paradigm. 
Implementing inter-class gradient matching can stabilize the training process of both positive and negative samples, thus seeking a balance between close set generalization and open set recognition. 

\vspace{0.3em}

\noindent \textbf{Remark.} The close set classifier can also be considered as a general form of multi-binary classifier, where each channel serves as a single-output binary classifier.
The introduction of the two-channel binary classifiers for open set classification is due to the fact that the cross entropy loss is not well-suited for single-output binary classifiers since it optimizes only the positive class but does not address the negative classes.

\begin{figure}[t]
\vspace{-0.1in}
\begin{algorithm}[H]
\caption{Training process of MEDIC++}  
\begin{algorithmic}[1]    
\label{alg:medic++}
\renewcommand{\algorithmicrequire}{\textbf{Input:}}
\REQUIRE Domains $\mathcal{S}$; classes $\mathcal{C}$;  split counts $t_d$ and $t_c$; tasks per step $m$; model parametrized by $\Theta$; loss function $\mathcal{L}$; learning rates $\alpha$ and $\epsilon$;
\WHILE{$\Theta$ not converged}
\STATE \textbf{Init} $\hat{\Theta} \leftarrow \Theta$;  $\mathcal{A} \leftarrow \varnothing$;
\STATE \textbf{Random split} $\mathcal{S}_1, \mathcal{S}_2, ..., \mathcal{S}_{t_d} \leftarrow \mathcal{S}$;
\FOR{$i=1, 2, ..., t_d$}
    \STATE \textbf{Random or adaptively split} $\mathcal{C}_1, \mathcal{C}_2, ..., \mathcal{C}_{t_c} \leftarrow \mathcal{C}$;
    \FOR{$j=1, 2, ..., t_c$}
        \STATE \textbf{Sample} ${\mathcal{B}}_{ij}$ from $(\mathcal{S}_i, \mathcal{C}_j)$;
        \STATE $\mathcal{A} \leftarrow \mathcal{A} \cup \{ {\mathcal{B}}_{ij} \}$;
    \ENDFOR
\ENDFOR
\WHILE{$\mathcal{A} \neq \varnothing$}
\STATE $\mathcal{B} \leftarrow$ random pop $m$ tasks from $\mathcal{A}$;
\STATE $\hat{\Theta} \leftarrow \hat{\Theta} - \alpha \cdot \nabla_{\hat{\Theta}} \mathcal{L}(\mathcal{B}; \hat{\Theta})$;
\ENDWHILE
\STATE $\Theta \leftarrow \Theta + \epsilon(\hat{\Theta}-\Theta)$;
\ENDWHILE
\end{algorithmic}
\end{algorithm}
\vspace{-0.22in}
\end{figure}

\subsection{Inference}
\label{subsec:method-inference}

In the test phase, each target sample is first predicted by the close set classifier to obtain a probability distribution $p(\hat{y}|x)$ over known classes.
The model either (i) chooses the value of its maximum likelihood:
\begin{equation}
\label{eq:confcls}
{\rm conf}_{\rm cls}(x) = {\rm max}_{i=1}^{|\mathcal{C}|}(p(\hat{y}|x)_i),
\end{equation}
or (ii) then refers to the corresponding one-vs-all classifier and chooses the value on its positive output channel as the confidence score \cite{saito2021ovanet}: 
\begin{equation}
\label{eq:confblcs}
{\rm conf}_{\rm bcls}(x) = p(\hat{y}^{{\rm argmax}_{i=1}^{|\mathcal{C}|}(p(\hat{y}|x)_i)}|x).
\end{equation}
If the score is greater than a preset threshold $\mu$, then classify the sample into known classes, otherwise judge it as unknown. 
Experimental results for these two inference modes are both reported in \cref{sec:experiment}.

\section{Theoretical Analysis}
\label{sec:theoretical}
\cref{subsec:appendix-a} proves that reptile-like meta-learning conducts pairwise gradient
matching between each step. 
\cref{subsec:appendix-b} explains how to scale the learning rate within the outer loop.
\cref{subsec:appendix-c} demonstrates that task-wise gradient matching is positively correlated with the step count.

\subsection{Step-wise Gradient Matching}
\label{subsec:appendix-a}
We prove \cref{eq:reptile-objective} through mathematical induction. Let's start by revisiting the definitions of the $n$ steps inner loop, during which the model's parameters transition from $\Theta$ to $\hat{\Theta}$. 
We represent the loss at each step as $\lbrace \mathcal{L}_1, \mathcal{L}_2, ..., \mathcal{L}_n\rbrace$, 
and the parameter updating trajectory as $\lbrace \theta_1, \theta_2, ..., \theta_{n+1}\rbrace$, with $\theta_1$ and $\theta_{n+1}$ corresponding to $\Theta$ and $\hat{\Theta}$ respectively.
We use $\mathcal{L}_i(\theta_j)$ to denote the loss of the $i$-th step on parameters $\theta_j$. 
During the inner loop, the update process is performed with a small learning rate $\alpha$: 
\begin{equation}
\label{eq:update-process}
\begin{split}
\theta_2 &= \theta_1 - \alpha\mathcal{L}_1'(\theta_1) \\
\theta_3 &= \theta_2 - \alpha\mathcal{L}_2'(\theta_2) \\
&\vdots  \\
\theta_{n+1} &= \theta_n - \alpha\mathcal{L}_n'(\theta_n).
\end{split}
\end{equation} 
While in the outer loop, the model is updated from $\theta_1$ to the final parameters:
\begin{equation}
\label{eq:theta-final}
\Theta \leftarrow \theta_1-\epsilon(\theta_1-\theta_{n+1}).
\end{equation}
We then sum the formulas from \cref{eq:update-process} and obtain:
\begin{equation}
\label{eq:update-process-sum}
\theta_1-\theta_{n+1} = \alpha\sum_{i=1}^n\mathcal{L}_i'(\theta_i).
\end{equation}
Plugging \cref{eq:update-process-sum} into \cref{eq:theta-final} yields the original representation for the reptile-like meta-learning objective:
\begin{equation}
\label{eq:reptile-objective-ori}
\mathop{\rm argmin}_{\theta_1}\, 
\alpha \sum_{i=1}^n\mathcal{L}_i(\theta_i).
\end{equation}

\noindent \textbf{Objective.} 
To prove that the gradient of any step is matched with those of the other $n$-1 steps, it is adequate to demonstrate that for any positive integer $i=k$, step $k$ is gradient-matched with the previous $k$-1 steps as:
\begin{equation}
\label{eq:lossk-thetak}
\mathcal{L}_k(\theta_k) = \mathcal{L}_k(\theta_1) - \alpha\sum_{i=1}^{k-1}\mathcal{L}_i'(\theta_1) \cdot \mathcal{L}_k'(\theta_1) + \mathcal{O}(\alpha^2),
\end{equation}
where gradient matching between two steps can be expressed by their dot product at $\theta_1$.
Proving \cref{eq:lossk-thetak} only requires that the following equation holds for any loss function $\mathcal{L}$:
\begin{equation}
\label{eq:loss-thetak}
\mathcal{L}(\theta_k) = \mathcal{L}(\theta_1) - \alpha\sum_{i=1}^{k-1}\mathcal{L}_i'(\theta_1) \cdot \mathcal{L}'(\theta_1) + \mathcal{O}(\alpha^2).
\end{equation}

\noindent \textbf{Base Case.} When $i$ equals 1, it is evident that $\mathcal{L}(\theta_i) = \mathcal{L}(\theta_1)$, so \cref{eq:loss-thetak} holds. When $i$ equals 2, we can substitute \cref{eq:update-process} into $\mathcal{L}(\theta_2)$ and conduct a first order Taylor expansion on it: 
\begin{equation}
\label{eq:loss-theta2}
\mathcal{L}(\theta_2) = \mathcal{L}(\theta_1) - \alpha\mathcal{L}_1'(\theta_1) \cdot \mathcal{L}'(\theta_1) + \mathcal{O}(\alpha^2),
\end{equation}
thus \cref{eq:loss-thetak} also holds.

\vspace{0.3em}

\noindent \textbf{Inductive Step.} Assuming that \cref{eq:loss-thetak} is true for arbitrary $i \leq k$, we prove its validity when $i$ equals $k+1$. 
Plugging \cref{eq:update-process} and \cref{eq:loss-thetak} into $\mathcal{L}(\theta_{k+1})$ yields:
\begin{equation}
\label{eq:loss-thetak+1}
\begin{split}
\mathcal{L}(\theta_{k+1}) =\ &\mathcal{L}(\theta_k) - \alpha\mathcal{L}_k'(\theta_k) \cdot \mathcal{L}'(\theta_k) + \mathcal{O}(\alpha^2) \\
=\ &\mathcal{L}(\theta_1) - \alpha\sum_{i=1}^{k-1}\mathcal{L}_i'(\theta_1) \cdot \mathcal{L}'(\theta_1) + \mathcal{O}(\alpha^2) \\
&- \alpha(\mathcal{L}_k'(\theta_1) + \mathcal{O}(\alpha))(\mathcal{L}'(\theta_1) + \mathcal{O}(\alpha)) + \mathcal{O}(\alpha^2) \\
=\ &\mathcal{L}(\theta_1) - \alpha\sum_{i=1}^{k}\mathcal{L}_i'(\theta_1) \cdot \mathcal{L}'(\theta_1) + \mathcal{O}(\alpha^2).
\end{split}
\end{equation}
Note that we substitute $\mathcal{L}_k'(\theta_k)$ into \cref{eq:loss-thetak} to obtain: 
\begin{equation}
\label{eq:----}
\mathcal{L}_k'(\theta_k) = \mathcal{L}_k'(\theta_1) - \alpha\sum_{i=1}^{k-1}\mathcal{L}_i'(\theta_1) \cdot \mathcal{L}_k''(\theta_1) + \mathcal{O}(\alpha^2),
\end{equation}
which is simplified as $\mathcal{L}_k'(\theta_1) + \mathcal{O}(\alpha)$ within \cref{eq:loss-thetak+1}.
The derivation of $\mathcal{L}'(\theta_k)$ follows the same process.

\vspace{0.3em}

\noindent \textbf{Conclusion.} We prove that \cref{eq:loss-thetak} holds for all positive integers $i=k$ and any loss function $\mathcal{L}$.
Plugging \cref{eq:loss-thetak} into \cref{eq:reptile-objective-ori} and the meta-objective is transformed to: 
\begin{equation}
\label{eq:reptile-objective-gamma}
\mathop{\rm argmin}_{\theta_1}\, \sum_{i=1}^{n}\mathcal{L}_i(\theta_1)-\alpha\sum_{i, j \in \mathcal{N}}^{i\neq j}\mathcal{L}_i'(\theta_1) \cdot \mathcal{L}_j'(\theta_1).
\end{equation}
By replacing the learning rate $\alpha$ with $\gamma$ and initial parameters $\theta_1$ with $\Theta$, we ultimately obtain \cref{eq:reptile-objective}.

\vspace{0.3em}

\noindent \change{\textbf{Remark.}
Our analysis is originally motivated by MLDG \cite{li2018learning}, which adopts a two-step procedure and derives gradient matching between two domain task splits. This observation strongly inspire MEDIC \cite{wang2023generalizable} and lead us to investigate whether such behavior could be extended to multi-step settings. 
The background explains why our derivation differs substantially from previous Reptile-based analyses \cite{nichol2018first, shi2021gradient, lee2022sequential}. Prior works formulate the objective in terms of mathematical expectation, specifically through a quantity referred as AvgGradInner \cite{nichol2018first}. They first derive the gradient at step $i$ as:
\begin{equation}
\label{eq: gradient-step-i}
\mathcal{L}'_i(\theta_i) = \mathcal{L}'_i(\theta_1) - \alpha \mathcal{L}''_i(\theta_1) \sum_{j=1}^{i-1}\mathcal{L}'_j(\theta_1) + \mathcal{O}(\alpha^2),
\end{equation}
and then isolate a single term from the second component and compute its expectation:
\begin{equation}
\label{eq: avggradinner}
\begin{aligned}
\mathop{\rm AvgGradInner}
&= \mathbb{E}_{i,j}(\mathcal{L}''_i(\theta_1) \mathcal{L}'_j(\theta_1)) \\
&= \mathbb{E}_{i,j}(\mathcal{L}''_j(\theta_1) \mathcal{L}'_i(\theta_1)) \\
&= \frac{1}{2}\mathbb{E}_{i,j}(\mathcal{L}''_i(\theta_1) \mathcal{L}'_j(\theta_1) 
+ \mathcal{L}''_j(\theta_1) \mathcal{L}'_i(\theta_1)) \\
&= \frac{1}{2}\mathbb{E}_{i,j}((\mathcal{L}'_i(\theta_1) 
\mathcal{L}'_j(\theta_1))')
\end{aligned}
\end{equation}
The result is therefore expressed in terms of an expectation $\mathbb{E}$, which cannot be eliminated from their formulation.
In contrast, we establish \cref{eq:lossk-thetak} 
which directly yields:
\begin{equation}
\label{eq: gradinner}
\mathcal{L}'_i(\theta_i) = \mathcal{L}'_i(\theta_1) - \alpha\sum_{j=1}^{i-1}(\mathcal{L}_j'(\theta_1) \cdot \mathcal{L}_i'(\theta_1))' + \mathcal{O}(\alpha^2).
\end{equation}
By removing the operator $\mathbb{E}$, we characterize the exact pairwise gradient matching among tasks, rather than its expectation. This distinction is central to our theoretical development, as it ensures that gradient matching holds on the current path instead of merely on average.}

\subsection{Scaling the Learning Rate}
\label{subsec:appendix-b}
We discuss how to set learning rate $\epsilon$ for the outer loop. 
Plugging \cref{eq:step-loss} into \cref{eq:reptile-objective-ori} leads to the loss function as:
\begin{equation}
\label{eq:-----}
\mathcal{L}_{\rm outer} = \alpha \sum_{i=1}^n \sum_{k=1}^{m_i}\mathcal{H}_{a_i^k}(\theta_i),
\end{equation}
while the standard loss of empirical risk minimization (ERM) \cite{naumovich1998statistical} without meta-learning can be expressed as:
\begin{equation}
\label{eq:------}
\begin{split}
\mathcal{L}_{\rm erm} &= \frac{1}{t} \sum_{i=1}^t \mathcal{H}_{i}(\theta_1) \\
&=  \frac{1}{t} \sum_{i=1}^n \sum_{k=1}^{m_i}\mathcal{H}_{i}(\theta_1),
\end{split}
\end{equation}
which implies that the coefficient of the loss in the outer loop is $\alpha t$ times that of ERM, thus $\epsilon$ needs to be scaled to $\frac{1}{\alpha t}$ of the default learning rate.

\begin{figure}[t]
    \centering
    \vspace{0.05in}
    \includegraphics[width=0.98\linewidth]{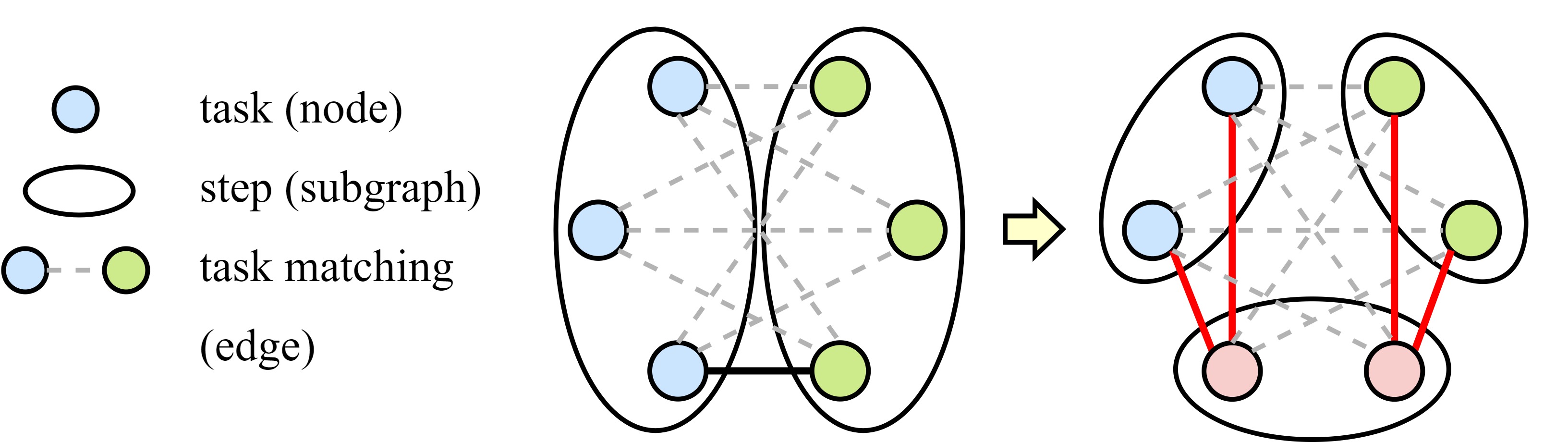}
    \caption{Supplementary visual aids for positive correlation between task-wise gradient matching and step count of \cref{subsec:appendix-c}.}
    \vspace{-0.1in}
    \label{fig:graph}
\end{figure}

\subsection{Relationship Between $t_n$ and $n$}
\label{subsec:appendix-c}

\change{
In \cref{subsec:method-medic++}, the estimate of the gradient matching count $t_n$ increasing with the step count $n$ is presented in continuous form. Here, we provide a discrete proof.
As shown in \cref{fig:graph}, we construct a graph model where tasks are represented as nodes, task-wise gradient matching as edges, and steps as node partitions $\lbrace \mathcal{P}_1, \mathcal{P}_2, ..., \mathcal{P}_n \rbrace$.
The number of tasks for the $i$-th step can also be written as the number of nodes $|\mathcal{P}_i|$.
Assuming a uniform task count distribution per step, it holds for any $i$ and $j$ that:
\begin{equation}
\label{eq:uniform-distribution}
|\mathcal{P}_i| \leq |\mathcal{P}_j| + 1.
\end{equation}
Matching gradient between steps $i$ and $j$ can be regarded as generating a complete bipartite graph $K(\mathcal{P}_i, \mathcal{P}_j)$ with a total of $|\mathcal{P}_i| \cdot |\mathcal{P}_j|$ edges.
When the number of steps transforms from $n+1$ to $n$, it is equivalent to uniformly dividing $\mathcal{P}_{n+1}$ into $n$ sub-partitions $\lbrace \mathcal{P}_1', \mathcal{P}_2', ..., \mathcal{P}_n' \rbrace$, and allocating them to the remaining partitions according to their respective indices. 
The change in the number of edges is expressed as:
\begin{equation}
\label{eq:delta-edge}
\begin{split}
\Delta_{n+1 \rightarrow n} &= \sum_{i, j \in \mathcal{N}}^{i \neq j}|\mathcal{P}_i'| \cdot |\mathcal{P}_j'| - \sum_{i=1}^n|\mathcal{P}_i'| \cdot |\mathcal{P}_i| \\
&= \frac{1}{2}\sum_{i=1}^n| \mathcal{P}_i'| \sum_{j \in \mathcal{N}}^{j \neq i}|\mathcal{P}_j'| - \sum_{i=1}^n|\mathcal{P}_i'| \cdot |\mathcal{P}_i| \\
&= \frac{1}{2} \sum_{i=1}^n|\mathcal{P}_i'| \left( \sum_{j \in \mathcal{N}}^{j \neq i}|\mathcal{P}_j'| - 2|\mathcal{P}_i|\right).
\end{split}
\end{equation}
The first term is the edges unique to step $n$, while the second term is the edges unique to step $n+1$. It is evident that each term of the summation in \cref{eq:delta-edge} equals $0$ when $|\mathcal{P}_i'|=0$,
For any $|\mathcal{P}_i'| \geq 1$, it follows from \cref{eq:uniform-distribution} that:
\begin{equation}
\label{eq:-------}
\sum_{j \in \mathcal{N}}^{j \neq i}|\mathcal{P}_j'| = |\mathcal{P}_{n+1}| - |\mathcal{P}_i'| \leq |\mathcal{P}_{n+1}| - 1 \leq |\mathcal{P}_{i}|,
\end{equation}
thus the corresponding terms are less than $0$. 
Because of this, the value of \cref{eq:delta-edge} is negative, so the number of edges from step $n+1$ to $n$ decreases, which shows a positive correlation between gradient matching count $t_n$ and step count $n$.}

\section{Experiment}
\label{sec:experiment}

\subsection{Datasets}
\label{subsec:experiment:datasets}

We evaluate on seven standard DG datasets whose details are described as follows:
(i) \textbf{PACS} \cite{li2017deeper} contains 4 domains (\emph{photo}, \emph{art-painting}, \emph{cartoon}, \emph{sketch}) with 7 classes and 9,991 images. 
(ii) \textbf{Office-Home} \cite{venkateswara2017deep} comprises 4 domains (\emph{art}, \emph{clipart}, \emph{product}, \emph{real-world}) with 65 classes and 15,588 images.
(iii) \textbf{VLCS} \cite{fang2013unbiased} consists of 4 domains (\emph{pascal}, \emph{labelme}, \emph{caltech}, \emph{sun}) with 5 classes and 10,729 images.
(iv) \textbf{TerraIncognita} \cite{beery2018recognition} is composed of 4 domains (\emph{location38}, \emph{location43}, \emph{location46}, \emph{location100}) with 100 classes and 24,788 images.
(v) \textbf{DomainNet} \cite{peng2019moment} includes 6 domains (\emph{clipart}, \emph{infograph}, \emph{painting}, \emph{quickdraw}, \emph{real}, \emph{sketch}) with 345 classes and 586,575 images.
(vi) \textbf{Digits-DG} \cite{zhou2020deep} is an aggregation of 4 domains (\emph{mnist} \cite{lecun1998gradient}, \emph{mnist-m} \cite{ganin2015unsupervised}, \emph{svhn} \cite{netzer2011reading}, \emph{syn} \cite{ganin2015unsupervised}) with 10 classes and 48,000 images.
(vii) \textbf{CMNIST} \cite{arjovsky2020invariant} consists of 3 domains, 2 classes, and 60,000 images.

\begin{table}[t]
\caption{Close set accuracy (\%) on DomainBed Benchmark.}
\vspace{-0.05in}
\centering
\begin{threeparttable}
\resizebox{1\linewidth}{!}{
\changetable
\begin{tabular}{lccccccc}
\toprule
\textbf{Method} & {\textbf{CMST}} & {\textbf{PACS}} & {\textbf{VLCS}} & {\textbf{Office}} & {\textbf{Terra}} & \textbf{DomNet} & {\textbf{Avg}} \\
\midrule
ERM \cite{naumovich1998statistical} & 51.5 & 85.5 & 77.5 & 66.5 & 46.1 & 40.9 & 61.3 \\
RSC \cite{huang2020self} & 51.7 & 85.2 & 77.1 & 65.5 & 46.6 & 38.9 & 60.8 \\
GroupDRO \cite{sagawa2019distributionally} & 52.1 & 84.4 & 76.7 & 66.0 & 43.2 & 33.3 & 59.3 \\
MLDG \cite{li2018learning} & 51.5 & 84.9 & 77.2 & 66.8 & 47.8 & 41.2 & 61.6 \\
V-REx \cite{krueger2021out} & 51.8 & 84.9 & 78.3 & 66.4 & 46.4 & 33.6 & 60.2 \\
CORAL \cite{sun2016deep} & 51.5 & 86.2 & \underline{78.8} & 68.7 & 47.7 & 41.5 & 62.4 \\
AND-mask \cite{parascandolo2020learning} & 51.3 & 84.4 & 78.1 & 65.6 & 44.6 & 37.2 & 60.2 \\
SAND-mask \cite{shahtalebi2021sand} & 51.8 & 84.6 & 77.4 & 65.8 & 42.9& 32.1 & 59.1 \\
Fish \cite{shi2021gradient} & 51.6 & 85.5 & 77.8 & 68.6 & 45.1 & \underline{42.7} & 61.9 \\
Fishr \cite{rame2022fishr} & 52.0 & 85.5 & 77.8 & 67.8 & 47.4 & 41.7 & 62.0 \\
HGP \cite{hemati2023understanding} & 51.8 & 84.7 & 77.6 & 68.2 & 43.6 & 41.1 & 61.2 \\ 
Hutchinson \cite{hemati2023understanding} & \underline{52.3} & 83.9 & 76.8 & 68.2 & 46.6 & 41.6 & 61.6  \\

\midrule

MEDIC++ \tnote{1} & 52.2 & \underline{87.3} & 78.5 & \underline{69.6} & \underline{49.3} & 40.7 & \underline{62.9}
 \\

MEDIC++ \tnote{2} & \textbf{52.4} & \textbf{89.4} & \textbf{79.0} & \textbf{71.6} & \textbf{50.2} & \textbf{46.7} & \textbf{64.9} \\
\bottomrule
\end{tabular}
}
\begin{tablenotes}
    \item[1] Using default model structure and hyperparameters.
    \item[2] Using our own hyperparameters with multi-binary classifier.
    \item[3] The best and second-best results are \textbf{bolded} and \underline{underlined} respectively.
\end{tablenotes}
\end{threeparttable}
\vspace{-0.1in}
\label{tab:domainbed-close}
\end{table}
\begin{table*}[t]
\caption{Results (\%) of PACS on ResNet50. (known: unknown = 6:1)}
\vspace{-0.05in}
\centering
\begin{threeparttable}
\resizebox{1.\linewidth}{!}{
\begin{tabular}{l|ccc|ccc|ccc|ccc|ccc}
\toprule
& \multicolumn{3}{c|}{\textbf{Photo}} & \multicolumn{3}{c|}{\textbf{Art}} & \multicolumn{3}{c|}{\textbf{Cartoon}} & \multicolumn{3}{c|}{\textbf{Sketch}} & \multicolumn{3}{c}{\textbf{Avg}} \\
\textbf{Method} & Acc & H-score & OSCR & Acc & H-score & OSCR & Acc & H-score & OSCR & Acc & H-score & OSCR & Acc & H-score & OSCR \\

\midrule
OpenMax \cite{bendale2016towards} & 97.58 & 93.09 & - & 88.37 & 73.91 & - & 84.38 & 68.23 & - & 80.07 & 68.06 & - & 87.60 & 75.82 & - \\

ARPL \cite{chen2021adversarial} & 97.09 & \textbf{96.81} & 96.86 & 88.24 & 77.48 & 80.32 & 82.68 & 67.19 & 68.31 & 78.08 & 70.04 & 69.47 & 86.52 & 77.88 & 78.74\\


MLDG \cite{li2018learning} & 96.77 & 95.85 & 96.33 & 87.99 & 77.16 & 79.93 & 83.45 & 68.74 & 71.32 & 82.25 & 73.16 & 72.27 & 87.61 & 78.73 & 79.96\\

ERM \cite{naumovich1998statistical} & 97.09 & 96.58 & 96.68 & 89.99 & 76.05 & 82.44 & 85.10 & 65.79 & 70.59 & 80.31 & 70.29 & 70.16 & 88.12 & 77.18 & 79.97 \\

Fish \cite{shi2021gradient} & 97.01 & 95.27 & 96.37 & 88.31 & 76.85 & 79.19 & 84.59 & 64.90 & 72.01 & 83.76 & 72.10 & 73.10 & 88.42 & 77.28 & 80.17 \\

CIRL \cite{lv2022causality} & 96.53 & 87.75 & 95.40 & 92.06 & 70.75 & 77.44 & 85.71 & 68.82 & 73.71 & 84.35 & 66.73 & 77.24 & 89.66 & 73.51 & 80.95 \\

MixStyle \cite{zhou2020domain} & 96.53 & 93.57 & 95.30 & 90.87 & 79.15 & 83.27 & 86.80 & 68.08 & 74.68 & 84.88 & 71.57 & 73.41 & 89.77 & 78.09 & 81.66\\

CrossMatch \cite{zhu2021crossmatch} & 96.53 & 96.34 & 96.12 & 91.37 & 75.67 & 82.32 & 83.92 & 67.02 & 74.55 & 81.61 & 72.03 & 73.99  & 88.37 & 77.76 & 81.75 \\

SWAD \cite{cha2021swad} & 96.37 & 84.56 & 93.24 & \textbf{93.75} & 68.41 & 85.00 & 85.57 & 58.57 & 75.90 & 81.90 & 74.66 & 74.65 & 89.40 & 71.55 & 82.20 \\

MVDG \cite{zhang2022mvdg} & 97.17 & 95.02 & 96.63 & 92.50 & 79.47 & 85.02 & 86.02 & 71.05 & 76.03 & 83.44 & 75.24 & 75.18 & 89.78 & 80.20 & 83.21 \\

\midrule

MEDIC & 96.37 & 94.75 & 95.79 & 91.62 & 81.61 & \textbf{85.81} & 86.65 & \textbf{77.39} & 78.30 & 84.61 & 78.35 & 79.50 & 89.81 & 83.03 & 84.85 \\

MEDIC++ & \textbf{97.58} & 96.56 & \textbf{96.99} & 93.25 & \textbf{82.70} & 85.75 & \textbf{87.58} & 76.57 & \textbf{78.43} & \textbf{85.98} & \textbf{78.36} & \textbf{79.63} & \textbf{91.10} & \textbf{83.55} & \textbf{85.20} \\

\bottomrule
\end{tabular}
}
\end{threeparttable}
\vspace{-0.075in}
\label{tab: pacs6-res50}
\end{table*}
\begin{table*}[t]
\caption{Results (\%) of Digits-DG on ConvNet. (known: unknown = 6:4)} 
\vspace{-0.05in}
\centering
\begin{threeparttable}

\resizebox{1.\linewidth}{!}{
\begin{tabular}{l|ccc|ccc|ccc|ccc|ccc}
\toprule
& \multicolumn{3}{c|}{\textbf{MNIST}} & \multicolumn{3}{c|}{\textbf{MNIST-M}} & \multicolumn{3}{c|}{\textbf{SVHN}} & \multicolumn{3}{c|}{\textbf{SYN}} & \multicolumn{3}{c}{\textbf{Avg}} \\
\textbf{Method} & Acc & H-score & OSCR & Acc & H-score & OSCR & Acc & H-score & OSCR & Acc & H-score & OSCR & Acc & H-score & OSCR \\

\midrule
OpenMax \cite{bendale2016towards} & 97.33 & 52.03 & - & 71.03 & 57.26 & - & 72.00 & 49.46 & - & 84.83 & 54.78 & - & 81.30 & 53.38 & - \\ 

MixStyle \cite{zhou2020domain} & 97.86 & 73.25 & 89.36 & \textbf{74.50} & 59.30 & 56.95 & 69.28 & 53.24 & 48.43 & 85.06 & 60.22 & 65.44 & 81.68 & 61.50 & 65.05\\

ERM \cite{naumovich1998statistical} & 97.47 & 80.90 & 92.60 & 71.03 & 53.92 & 54.04 & 71.08 & 54.37 & 49.86 & 85.67 & 51.57 & 67.63 & 81.31 & 60.19 & 66.03\\

ARPL \cite{chen2021adversarial} & 97.75 & \textbf{85.74} & 91.86 & 69.78 & 58.08 & 54.21 & 71.78 & 56.98 & 53.63 & 85.31 & 64.04 & 65.89 & 81.16 & 66.21 & 66.40\\

MLDG \cite{li2018learning} & 97.83 & 80.36 & 94.28 & 71.11 & 46.84 & 55.17 & 73.64 & 53.54 & 53.64 & 86.08 & 63.56 & 70.34 & 82.16 & 61.08 & 68.36 \\

SWAD \cite{cha2021swad} & 97.71 & 84.44 & 92.65 & 73.09 & 53.35 & 55.94 & 76.08 & 59.18 & 56.25 & 87.95 & 51.27 & 69.03 & 83.71 & 62.06 & 68.47 \\

Fish \cite{shi2021gradient} & 97.83 & 74.69 & 95.61 & 71.81 & 47.31 & 56.05 & 74.42 & 49.94 & 52.89 & 86.14 & 66.57 & 73.88 & 82.55 & 59.63 & 69.61 \\

CIRL \cite{lv2022causality} & 97.92 & 81.14 & 93.50 & 73.78 & 59.88 & 58.25 & \textbf{80.06} & 58.73 & 56.88 & 87.86 & 64.91 & 69.95 & \textbf{84.91} & 66.16 & 69.64 \\

\midrule

MEDIC & 97.89 & 83.20 & 95.81 & 71.14 & \textbf{60.98} & 58.28 & 76.00 & 58.77 & 57.60 & \textbf{88.11} & 62.24 & 72.91 & 83.28 & 66.30 & 71.15\\

MEDIC++ & \textbf{98.44} & 80.58 & \textbf{96.88} & 73.14 & 60.51 & \textbf{61.62} & 77.31 & \textbf{61.55} & \textbf{59.90} & 87.43 & \textbf{67.67} & \textbf{76.65} & 84.08 & \textbf{67.58} & \textbf{73.76} \\

\bottomrule
\end{tabular}
}
\end{threeparttable}
\vspace{-0.075in}  
\label{tab: digits6-conv}
\end{table*}
\begin{table*}[t]
\caption{Results (\%) of PACS on GFNet-H-Ti. (known: unknown = 6:1)}
\vspace{-0.05in}
\centering
\begin{threeparttable}
\resizebox{1.\linewidth}{!}{
\begin{tabular}{l|ccc|ccc|ccc|ccc|ccc}
\toprule
& \multicolumn{3}{c|}{\textbf{Photo}} & \multicolumn{3}{c|}{\textbf{Art}} & \multicolumn{3}{c|}{\textbf{Cartoon}} & \multicolumn{3}{c|}{\textbf{Sketch}} & \multicolumn{3}{c}{\textbf{Avg}} \\
\textbf{Method} & Acc & H-score & OSCR & Acc & H-score & OSCR & Acc & H-score & OSCR & Acc & H-score & OSCR & Acc & H-score & OSCR \\

\midrule
OneRing \cite{yang2022one} & 96.20 & 84.95 & - & 89.24 & 71.54 & - & 85.36 & 64.53 & - & 82.28 & 63.97 & - & 88.27 & 71.25 & -   \\

CrossMatch \cite{zhu2021crossmatch} & 96.93 & 81.83 & 88.86 & 91.56 & 74.25 & 80.69 & 85.15 & 69.48 & 70.00 & 82.49 & 68.66 & 72.39 & 89.03 & 73.56 & 77.98 \\

ALOFT \cite{guo2023aloft} & 97.90 & 86.45 & 89.98 & 93.75 & 74.12 & 79.89 & 85.98 & 68.80 & 71.15 & 83.58 & 67.30 & 72.41 & 90.30 & 74.17 & 78.36 \\

Fish \cite{shi2021gradient} & 97.66 & 87.17 & 88.95 & 91.24 & 76.75 & 81.29 & 85.62 & 71.50 & \textbf{74.89} & 84.48 & 72.45 & 74.18 & 89.75 & 76.97 & 79.83 \\

ERM \cite{naumovich1998statistical} & 97.58 & 91.23 & \textbf{95.87} & 91.06 & 74.69 & 82.64 & 85.00 & 62.78 & 70.07 & 81.88 & 73.23 & 73.05 & 88.88 & 75.48 & 80.41 \\

MLDG \cite{li2018learning} & 96.93 & 84.58 & 93.40 & 91.68 & 78.67 & 81.45 & 85.10 & 71.76 & 74.93 & 82.22 & 76.62 & 73.54 & 88.98 & 77.91 & 80.83 \\

ARPL \cite{chen2021adversarial} & 97.98 & \textbf{95.31} & 96.44 & 93.37 & 81.05 & 84.83 & 85.52 & 67.66 & 72.63 & 82.59 & 77.12 & 73.19 & 89.87 & 80.28 & 81.77  \\

SWAD \cite{cha2021swad} & 97.42 & 87.53 & 94.58 & 92.56 & 78.43 & 85.02 & \textbf{87.22} & 69.29 & 74.68 & 83.63 & 74.57 & 75.85 & 90.21 & 77.45 & 82.53\\

\midrule
MEDIC & 98.06 & 93.88 & 95.75 & 92.81 & 80.58 & 85.29 & 86.08 & 71.16 & 75.22 & 84.77 & 76.35 & 76.41 & 90.43 & 80.49 & 83.17 \\

MEDIC++ & \textbf{98.06} & 92.87 & 95.59 & \textbf{93.50} & \textbf{82.27} & \textbf{86.69} & 87.11 & \textbf{72.73} & 74.65 & \textbf{86.60} & \textbf{77.97} & \textbf{78.81} & \textbf{91.32} & \textbf{81.46} & \textbf{83.94} \\

\bottomrule
\end{tabular}
}

\end{threeparttable}
\vspace{-0.1in}
\label{tab: pacs6-gfnet}
\end{table*}

\subsection{Implementation Details}
\label{subsec:experiment:implementation}

\noindent \textbf{Basic details.} Each training set is randomly segmented into 3 parts by both domain and class to obtain a total of 9 tasks.
The inner loop comprises 3 steps, each of which contains 3 tasks with a fixed learning rate of $0.01$. 
For datasets other than Digit-DG, we employ ResNet50 \cite{he2016deep} and GFNet \cite{rao2021global} pretrained on ImageNet \cite{deng2009imagenet} as our backbone networks. 
For Digits-DG, we adopt a lightweight convolutional architecture called ConvNet from \cite{zhou2020deep}.
The leave-one-domain-out evaluation protocol is applied to all benchmarks, \emph{i.e.}, picking one target domain for testing and using the rest for training and validation. 
We set aside 20\% of the samples for validation from each source domain and choose the model that maximizes the accuracy on the overall validation set, which is same as the \emph{training-domain validation set} recommended in \cite{gulrajani2020search}. 

\noindent \change{\textbf{DomainBed benchmark.} We first follow the close set protocol proposed in \cite{gulrajani2020search}, including the hyperparameter search space, model structure and Adam optimizer. 
For datasets like PACS which contains 3 training domains with a default batch size of $(32, 96)$, we configure our batch size to $(12, 108)$. 
For DomainNet which includes 5 training domains with a default batch size of $(32, 160)$, we select our batch size as $(18, 162)$. 
For CMNIST with a default batch size of $(64, 128)$, we set our batch size as $(32, 128)$.
The first value in the parentheses is the batch size per domain or per task, while the second value is their combined sum.
The default learning rates are $5\times 10^{-5}$ for non-digits and $1\times 10^{-3}$ for digits, with our rates set to $6\times 10^{-4} \times 0.09$ and $2\times 10^{-2} \times 0.04$. 
As used in \cite{cha2021swad}, we triple the number of iterations for DomainNet from 5,000 to 15,000 since 5,000 iterations is inadequate for convergence.
Notably, the multi-binary classifier is excluded in this benchmark.}
 
\noindent \textbf{Additional configurations.} To further improve the model's performance, we fine-tune hyperparameters and incorporate the multi-binary classifier into training.
The number of iterations is doubled to 30,000 for DomainNet and 10,000 for TerraIncognita, while the batch size per task is uniformly set to 16 across all datasets.
Using a SGD optimizer, the initial learning rates of the outer loop are configured as follows: $0.02$ for PACS and Office-Home, $0.01$ for VLCS, TerraIncognita and DomainNet, $0.1$ for Digits-DG and CMNIST.
Each learning rate is then reduced to 10\%  in the last 20\% iterations. 
For open set experiments, the classes are organized in alphabetical order, with the former part as known classes and the latter as unknown.
The split rates for known and unknown classes on each dataset are detailed in the corresponding table caption.
We use close set validation accuracy for model selection.

\change{
For ablation studies, all methods use the same number of samples per iteration and scaled learning rate. Except for the specific ablation involving multi-binary classifiers, all methods are implemented with them to ensure a fair comparison.}

\begin{table*}[t]
\caption{Results (\%) of DomainNet on ResNet50. (known: unknown = 100:245)}
\vspace{-0.05in}
\centering
\begin{threeparttable}
\resizebox{0.95\linewidth}{!}{
\begin{tabular}{l|cc|cc|cc|cc|cc|cc|cc}
\toprule
& \multicolumn{2}{c|}{\textbf{clip}} & \multicolumn{2}{c|}{\textbf{info}} & \multicolumn{2}{c|}{\textbf{paint}} & \multicolumn{2}{c|}{\textbf{quick}} & \multicolumn{2}{c|}{\textbf{real}} & \multicolumn{2}{c|}{\textbf{sketch}} & 
\multicolumn{2}{c}{\textbf{Avg}} \\
\textbf{Method} & Acc & OSCR & Acc & OSCR & Acc & OSCR & Acc & OSCR & Acc & OSCR & Acc & OSCR & Acc & OSCR \\

\midrule
MIRO \cite{cha2022domain} & 71.55 & 61.45 & 31.38 & 25.03 & 59.66 & 50.78 & 19.06 & 13.27 & \textbf{75.95} & 66.07 & 65.14 & 55.07 & 53.79 & 45.28 \\
MixStyle \cite{zhou2020domain} & 71.10 & 61.42 & 30.46 & 25.14 & 60.96 & 51.47 & 21.68 & 15.85 & 73.13 & 62.81 & 67.04 & 58.04 & 54.06 & 45.79 \\
ERM \cite{naumovich1998statistical} & 71.62 & 61.67 & 31.62 & 26.06 & 61.17 & 51.32 & 21.06 & 15.31 & 75.13 & 65.21 & 67.40 & 57.72 & 54.67 & 46.21\\
ARPL \cite{chen2021adversarial} & 73.25 & 63.52 & 30.05 & 23.92 & 62.65 & 52.12 & 22.10 & 15.95 & 73.70 & 63.39 & 68.23 & 58.95 & 55.00 & 46.31 \\
CrossMatch \cite{zhu2021crossmatch} & 72.15 & 62.09 & 32.38 & 26.19 & 61.40 & 51.61 & 21.47 & 15.75 & 74.73 & 65.06 & 67.79 & 58.19 & 54.99 & 46.48  \\
MLDG \cite{li2018learning} & 71.63 & 62.04 & 32.56 & 26.85 & 61.55 & 51.20 & 21.74 & 15.86 & 75.34 & 65.19 & 68.12 & 58.68 & 55.16 & 46.64\\
Fish \cite{shi2021gradient} & \textbf{73.39} & 63.71 & 31.79 & 26.42 & 62.29 & 52.41 & 21.11 & 15.21 & 74.49 & 64.40 & 68.01 & 58.33 & 55.18 & 46.75 \\ 
SWAD \cite{cha2021swad} & 72.83 & 62.98 & 31.87 & 26.63 & \textbf{63.40} & 53.89 & \textbf{23.40} & \textbf{17.30} & 75.14 & 65.08 & 68.39 & 59.25 & 55.84 & 47.52\\

\midrule

MEDIC & 73.31 & 64.05 & 31.32 & 26.13 & 63.15 & 53.96 & 22.45 & 16.64 & 75.87 & 65.89 & 68.50 & 59.64 & 55.77 & 47.72 \\

MEDIC++ & 73.38 & \textbf{64.07} & \textbf{32.93} & \textbf{27.37} & 63.07 & \textbf{54.12} & 22.48 & 16.67 & 75.65 & \textbf{66.18} & \textbf{69.29} & \textbf{59.91} & \textbf{56.13} & \textbf{48.05} \\

\bottomrule
\end{tabular}
}
\end{threeparttable}
\vspace{-0.1in}
\label{tab: domainnet100-res50}
\end{table*}
\subsection{Evaluation Metrics}
\label{subsec:experiment-metric}

We choose three evaluation metrics that take both known and unknown class accuracy into account:
(i) \textbf{Acc} represents the typical close set accuracy. 
(ii) \textbf{H-score} \cite{fu2020learning} quantifies the harmonic mean of known class accuracy ${\rm acc}_{\rm k}$ and unknown class accuracy ${\rm acc}_{\rm u}$
as follows: 
\begin{equation}
\label{eq:h-score}
{\rm H\mbox{-}score}=2 \cdot \frac{{\rm acc}_{\rm k}\cdot{\rm acc}_{\rm u}}{{\rm acc}_{\rm k}+{\rm acc}_{\rm u}}.
\end{equation}
When ${\rm acc}_{\rm k} + {\rm acc}_{\rm u}$ remains constant, the closer ${\rm acc}_{\rm k}$ and ${\rm acc}_{\rm u}$ are, the larger H-score will be.
Compared with the weighted average, H-score puts more emphasis on the balance between close set classification and open set recognition.
Nevertheless, the manually designed threshold to reject unknown classes is not applicable for a random target domain.
We propose a threshold-independent metric (iii) open set classification rate (\textbf{OSCR}) \cite{dhamija2018reducing} which plots the true positive rate against the false positive rate using an ever-moving threshold.
Different from area under the receiver operating characteristic (AUROC) \cite{neal2018open} that neglects
known class accuracy, OSCR considers only correctly labeled samples as true positive ones.

\subsection{Close Set Results}
\label{subsec:close}

We compare our strategy with closely related meta-learning methods such as MLDG \cite{li2018learning} and Fish \cite{shi2021gradient}, as well as gradient-based methods like RSC \cite{huang2020self}, AND-mask \cite{parascandolo2020learning} and Fishr \cite{rame2022fishr}. 
Firstly, we use the same model architecture and default hyperparameters.
As \cref{tab:domainbed-close} illustrates, our method achieves the highest average performance, outperforming other methods on three datasets namely PACS, OfficeHome and TerraIncognita, 
surpassing the second-best method by 1.1\%, 0.9\% and 1.5\% respectively.
Secondly, by involving the multi-binary classifier, the model's average performance is further boosted by 2.0\%, suggesting that the binary classifiers which separate inter-class samples also play a role in improving close set accuracy.

\subsection{Open Set Results}
\label{subsec:open}

We conduct open set experiments, and the results on PACS, Digits-DG, and DomainNet are shown in \cref{tab: pacs6-res50}, \cref{tab: digits6-conv}, \cref{tab: pacs6-gfnet}, and \cref{tab: domainnet100-res50}, respectively.
Our strategy outperforms other DG and OSR methods in both close set and open set scenarios.  MEDIC++ shows superior performance compared to MEDIC, achieving a improvement in close set accuracy by 1.29\% in \cref{tab: pacs6-res50}. These results indicate that our method is capable of producing generalizable and discriminative representations, benefiting both DG and OSR tasks.

We also compare with several open set recognition methods such as OpenMax \cite{bendale2016towards} and ARPL \cite{chen2021adversarial}.
Note that we exclude the calculation of OSCR for OpenMax \cite{bendale2016towards} and OneRing \cite{yang2022one} due to their threshold-independent property, \emph{i.e.}, the classifier is configured with $|\mathcal{C}| + 1$ output channels, one of which is dedicated to the probability of unknown classes.
However, the H-score of them is still below average, further highlighting that the hard inference derived from source domains is not suitable for the unseen target domains.
It can be observed that ARPL \cite{chen2021adversarial}, which is one of the state-of-the-art approaches for open set recognition, fails to perform well compared to the standard DG methods.
This may indicate that the deep learning models can exhibit a natural inclination to recognize unknown classes, so the close set classification under distribution shift remains crucial in open set domain generalization.

\begin{table}[t]
\caption{Ablation study (\%) of tasks per step on PACS / ResNet50.}
\vspace{-0.05in}
\centering
\resizebox{0.95\linewidth}{!}{
\begin{tabular}{c|ccccc}
\toprule
\textbf{Tasks / Step} & {\textbf{Photo}} & {\textbf{Art}} & {\textbf{Cartoon}} & {\textbf{Sketch}} & {\textbf{Avg}} \\
\midrule
\multicolumn{6}{c}{Acc} \\
\midrule {}
9 & 97.42 & 90.56 & 84.54 & 81.88 & 88.60 \\ {}
5 4 & 97.09 & 92.68 & 86.44 & 85.01 & 90.31 \\ {}
3 3 3 & \textbf{97.58} & \textbf{93.25} & \textbf{87.58} & \textbf{85.98} & \textbf{91.10} \\ {}
3 2 2 2 & 97.01 & 93.06 & 86.70 & 84.37 & 90.28 \\ {}
2 2 2 2 1 & 97.25 & 92.43 & 87.11 & 84.13 & 90.23 \\ {}
2 2 1 1 1 1 1 & 95.96 & 91.24 & 86.19 & 82.28 & 88.92 \\ {}
1 1 1 1 1 1 1 1 1 & 95.80 & 86.30 & 82.42 & 81.72 & 86.56 \\ 

\midrule
\multicolumn{6}{c}{OSCR} \\
\midrule {}

9 & \textbf{97.16} & 82.95 & 72.31 & 74.23 & 81.66 \\ {}
5 4 & 96.73 & 85.13 & 76.77 & 79.20 & 84.46 \\ {}
3 3 3 & 96.99 & \textbf{85.75} & \textbf{78.43} & \textbf{79.63} & \textbf{85.20} \\ {}
3 2 2 2 & 96.44 & 85.50 & 77.84 & 78.18 & 84.49 \\ {}
2 2 2 2 1 & 96.71 & 84.62 & 77.77 & 78.76 & 84.46 \\ {}
2 2 1 1 1 1 1 & 95.32 & 79.56 & 75.74 & 75.17 & 81.45 \\ {}
1 1 1 1 1 1 1 1 1 & 91.42 & 75.80 & 71.48 & 70.39 & 77.27 \\ 
\bottomrule
\end{tabular}
}
\vspace{-0.05in}
\label{tab: step-cmp}
\end{table}

\begin{table}[t]
\caption{
Ablation studies (\%) of parameter sharing on PACS / ResNet50.
}
\vspace{-0.05in}
\centering
\begin{threeparttable}
\resizebox{0.95\linewidth}{!}{
\begin{tabular}{lcccccc}
\toprule
\textbf{Method} & \emph{share} & \textbf{Photo} & \textbf{Art} & \textbf{Cartoon} & \textbf{Sketch} & \textbf{Avg} \\

\midrule

\multicolumn{7}{c}{Acc} \\

\midrule

\multirow{2}{*}{MEDIC} & - & 96.37 & 91.62 & 86.65 & 84.61 & 89.81  \\
 & \checkmark & 97.01 & 92.18 & 86.70 & 85.27 & 90.29 \\

\midrule

\multirow{2}{*}{MEDIC++} & - & 97.58 & 93.25 & 87.58 & 85.98 & 91.10  \\
 & \checkmark & 97.17 & 92.43 & 87.22 & 85.91 & 90.68\\

\midrule

\multicolumn{7}{c}{OSCR} \\

\midrule

\multirow{2}{*}{MEDIC} & - & 95.79 & 85.81 & 78.30 & 79.50 & 84.85 \\
 & \checkmark & 96.24 & 85.21 & 77.57 & 79.35 & 84.59\\

\midrule

\multirow{2}{*}{MEDIC++} & - & 96.99 & 85.75 & 78.43 & 79.63 & 85.20 \\
 & \checkmark & 96.72 & 84.94 & 78.56 & 78.88 & 84.78 \\

\bottomrule
\end{tabular}
} 
\end{threeparttable}
\vspace{-0.05in}
\label{tab: param-share}
\end{table}
\begin{table}[t]
\caption{
Ablation studies (\%) of classifiers on PACS / ResNet50.
}
\vspace{-0.05in}
\centering
\begin{threeparttable}
\resizebox{0.95\linewidth}{!}{
\begin{tabular}{lcccccccc}
\toprule
\textbf{Method} & Tr-b \tnote{1} & Inf-c \tnote{2} & Inf-b \tnote{3} & \textbf{P} & \textbf{A} & \textbf{C} & \textbf{S} & \textbf{Avg} \\









\midrule
\multicolumn{9}{c}{OSCR} \\
\midrule

& - & \checkmark & - & 96.7 & 82.4 & 70.6 & 70.2 & 80.0\\
ERM \cite{naumovich1998statistical} & \checkmark & \checkmark & - & \textbf{97.3} & 83.9 & 70.7 & 70.9 & 80.7\\
& \checkmark & - & \checkmark & 97.1 & 83.8 & 71.1 & 72.0 & 81.0\\

\midrule

& - & \checkmark & - & 96.3 & 79.9 & 71.3 & 72.3 & 80.0\\
MLDG\cite{li2018learning} & \checkmark & \checkmark & - & 96.7 & 83.1 & 74.6 & 73.5 & 82.0\\
& \checkmark & - & \checkmark & 96.8 & 83.3 & 75.4 & 74.3 & 82.5\\

\midrule
& - & \checkmark & - & 96.4 & 79.2 & 72.0 & 73.1 & 80.2 \\
Fish \cite{shi2021gradient} & \checkmark & \checkmark & - & 96.2 & 81.1 & 75.6 & 72.0 & 81.2 \\
& \checkmark & - & \checkmark & 96.1 & 81.2 & 76.2 & 73.1 & 81.7 \\

\midrule

& - & \checkmark & - & 95.1 & 83.7 & 73.7 & 75.5 & 82.0\\
MEDIC & \checkmark & \checkmark & - & 95.4 & 84.7 & 77.5 & 76.8 & 83.6\\
& \checkmark & - & \checkmark & 95.8 & \textbf{85.8} & 78.3 & 79.5 & 84.9\\

\midrule

& - & \checkmark & - & 96.7 & 84.8 & 74.2 & 73.7 & 82.4 \\
MEDIC++ & \checkmark & \checkmark & - & 96.7  & 84.9 & 77.2 & 77.0 & 84.0 \\
& \checkmark & - & \checkmark & 97.0 & 85.8 & \textbf{78.4} & \textbf{79.6} & \textbf{85.2} \\

\bottomrule
\end{tabular}
}
\begin{tablenotes}
    \item[1] Training with multi-binary classifier. 
    \item[2] Inference with close set classifier.
    \item[3] Inference with multi-binary classifier.
\end{tablenotes}
\end{threeparttable}
\vspace{-0.05in}
\label{tab: cls-cmp}
\end{table}

\subsection{Ablation Study}
\label{subsec:ablation}

\noindent \textbf{Varying the number of steps.} 
As shown in \cref{tab: step-cmp}, we investigate the influence of tasks per step on the performance of model. We observe that both accuracy and OSCR initially increase, but then decline as the number of steps rises. This indicates that when the step count is relatively smaller, the notable expansion of gradient-matched tasks contributes to a rapid improvement. However, in the later stages, the differences between classes across steps lead to high similarity in the normalized features of different classes, making it challenging for the model to distinguish between them. Note that conventional meta-learning-based domain generalization methods do not experience this phenomenon because tasks from different domains are normalized separately, which actually aids in the extraction of domain-invariant features. 
From the second to the fourth rows are all variants of MEDIC++, which is based on the core idea that the step matters, but not too many. The similar results actually reflect the robustness of our method. In contrast, existing methods typically fall into one of two extremes: either using a single step or restricting each step to a single task. These approaches either completely lack gradient matching or are constrained by batch normalization and cost. As illustrated in the first and last rows of the table, both of these strategies significantly underperform MEDIC++.

\vspace{0.3em}

\noindent \textbf{The effect of different domain generalization paradigms.}
We compare our method with the baseline ERM \cite{naumovich1998statistical}, as well as meta-learning paradigms MLDG \cite{li2018learning} and Fish \cite{shi2021gradient}.
Both MLDG and Fish share the same concept of simulating virtual target domains, but ignoring the relationship among classes.
When using the same loss function and model, the strategy becomes the only variable between our method and others.
As shown in \cref{tab: cls-cmp}, both MEDIC and MEDIC++ outperform above methods no matter which option is uniformly appointed, demonstrating the critical role played by dualistic gradient matching in open set recognition.
Moreover, after transitioning from the cross-entropy loss (\emph{i.e.}, training with close set classifier only) to open set loss function (\emph{i.e.}, training with the two classifiers), our method ushers in the largest performance gain on the average of OSCR by $2.9\%$ and $2.8\%$ for MEDIC and MEDIC++ respectively, indicating that our strategy has better compatibility with the multi-binary classifier to learn a more generalizable boundary for each known class. 

\vspace{0.3em}

\begin{figure}[tp]
    \centering  
    \vspace{0.1in}
    \includegraphics[width=0.9\linewidth]{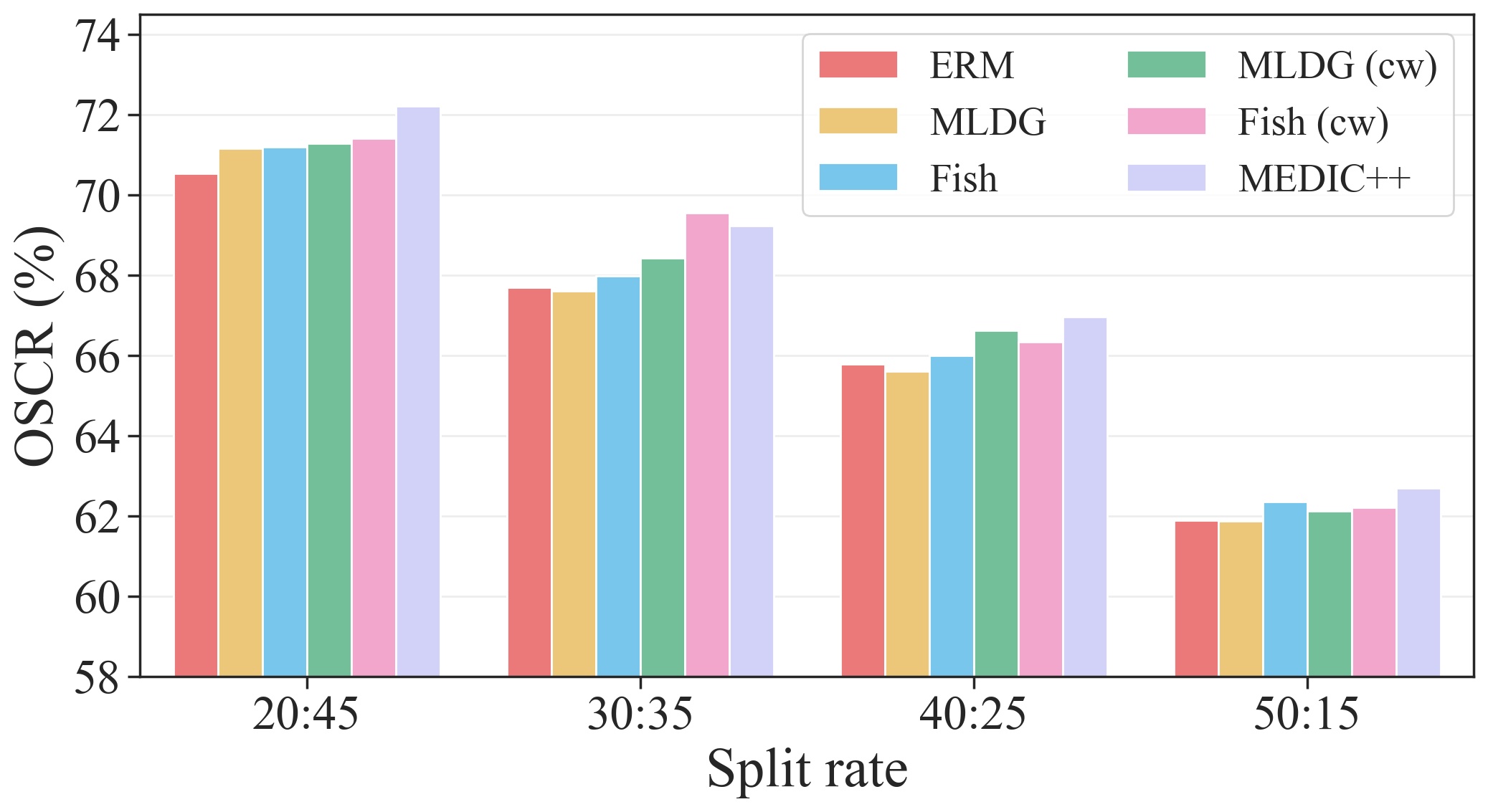}
    \vspace{-0.05in}
    \caption{The values of OSCR (\%) with the varying known-unknown splits on Office-Home using ResNet50, where the option \emph{cw} denotes the class-wise meta-learning version of corresponding strategy.}
    \vspace{-0.1in}
    \label{fig:office-range}
\end{figure}

\noindent \textbf{Varying the proportion of known to unknown.}
We conduct experiments on the Office-Home \cite{venkateswara2017deep} dataset using the multi-binary classifier across all strategies. The OSCR results are visually presented in \cref{fig:office-range}. It is evident that increasing the number of known classes introduces greater challenges to the classification task with lower accuracy. Notably, MEDIC++ achieves optimal performance across most split rates, which highlights the robustness of our method in diverse scenarios. We further construct class-wise meta-learning variants for each method. Interestingly, they consistently outperforms traditional domain-wise meta-learning strategies. In some cases, such as when the number of known classes is 30, it even surpasses MEDIC++. This thereby emphasizes the importance of finding optimal balance between classes, making it a crucial consideration in the design of meta-learning strategies.

\vspace{0.3em}

\noindent \textbf{On the effect of different meta-learning paradigms.} Ablation studies are also conducted to evaluate different optimization, task partitioning, and sampling techniques as detailed in \cref{tab: strategy-cmp}. 
The baseline is to split source data without considering domains or classes, gradually extending to domain-wise and class-wise partitions with added sampling strategies.
We find that both task partition methods benefit most optimization strategies except for MAML \cite{finn2017model}. This may be due to its dependence on only the final gradient during the inner loop, which is more random and thus requires more careful optimization.
The similarity between \cite{liu2020adaptive} and MEDIC++ is that both select confusable class pairs, while we further assign them to different steps rather than training together.
Compared to others, our strategy (\ie, the last line), which separates easily confusable class pairs across different steps, achieves the best open set performance. This suggests that enhancing gradient matching between these pairs improves the generalizability of unbiased decision boundaries.

\begin{table}[t]
\caption{Results (\%) of partial classes on PACS / ResNet50.}
\vspace{-0.05in}
\centering
\changetable
\begin{tabular}{l|ccccc}
\toprule
\textbf{Method} & {\textbf{Photo}} & {\textbf{Art}} & {\textbf{Cartoon}} & {\textbf{Sketch}} & {\textbf{Avg}} \\
\midrule
\multicolumn{6}{c}{Acc} \\
\midrule
ERM \cite{naumovich1998statistical} & 93.9 & 83.6 & 67.8 & 72.5 & 79.5 \\
MLDG \cite{li2018learning} & 94.6 & 89.4 & 70.9 & 78.4 & 83.3 \\
Fish \cite{shi2021gradient} & 96.0 & 85.8 & 71.5 & 80.3 & 83.4 \\
MEDIC & 95.7 & 89.4 & 72.0 & 81.2 & 84.6 \\
MEDIC++ & \textbf{96.3} & \textbf{89.8} & \textbf{73.3} & \textbf{82.4} & \textbf{85.4} \\
\midrule
\multicolumn{6}{c}{OSCR} \\
\midrule
ERM \cite{naumovich1998statistical} & 91.3 & 73.6 & 59.6 & 56.1 & 70.1 \\
MLDG \cite{li2018learning} & 93.3 & 78.2 & 62.8 & 71.2 & 76.4 \\
Fish \cite{shi2021gradient} & \textbf{95.3} & 74.5 & 63.3 & 71.6 & 76.2 \\
MEDIC & 94.7 & 79.5 & 64.5 & 71.2 & 77.5 \\
MEDIC++ & 95.1 & \textbf{80.7} & \textbf{66.0} & \textbf{71.8} & \textbf{78.4} \\
\bottomrule
\end{tabular}
\vspace{-0.05in}
\label{tab: pacs6-partial4}
\end{table}
\begin{table}[t]
\caption{Results (\%) of single domain on PACS / ResNet50.}
\vspace{-0.05in}
\centering
\changetable
\begin{tabular}{l|ccccc}
\toprule
\textbf{Method} & {\textbf{Photo}} & {\textbf{Art}} & {\textbf{Cartoon}} & {\textbf{Sketch}} & {\textbf{Avg}} \\
\midrule
\multicolumn{6}{c}{Acc} \\
\midrule
ERM \cite{naumovich1998statistical} & 45.6 & 68.5 & 73.4 & 48.7 & 59.0 \\ 
CM \cite{zhu2021crossmatch} & 47.0 & 67.5 & 75.1 & 58.7 & 62.1 \\ 
MEDIC & 54.3 & 74.7 & 81.1 & 69.6 & 69.9 \\ 
MEDIC++ & \textbf{60.6} & \textbf{75.4} & \textbf{82.0} & \textbf{70.8} & \textbf{72.2} \\ 

\midrule
\multicolumn{6}{c}{OSCR} \\
\midrule

ERM \cite{naumovich1998statistical} & 37.9 & 61.2 & 65.5 & 36.1 & 50.2 \\ 
CM \cite{zhu2021crossmatch} & 39.3 & 61.9 & 69.0 & 45.8 & 54.0 \\ 
MEDIC & 45.5 & 69.5 & 76.4 & 57.2 & 62.1 \\ 
MEDIC++ & \textbf{48.4} & \textbf{71.0} & \textbf{76.7} & \textbf{60.7} & \textbf{64.2} \\ 
\bottomrule
\end{tabular}
\vspace{-0.1in}
\label{tab: pacs6-single}
\end{table}

\begin{table}[t]
\caption{
Ablation studies (\%) of learning paradigms and \\ task sampling strategies on PACS / ResNet50.
}
\vspace{-0.05in}
\centering
\begin{threeparttable}
\resizebox{1\linewidth}{!}{
\changetable
\begin{tabular}{lccccccccc}
\toprule
\textbf{Baseline} & $dw$ \tnote{1} & $cw$ \tnote{2} & $opt$ \tnote{3} & \textbf{P} & \textbf{A} & \textbf{C} & \textbf{S} & \textbf{Avg} \\
\midrule
\multicolumn{9}{c}{Acc} \\
\midrule
\multirow{3}{*}{ERM \cite{naumovich1998statistical}} & - & - & (i) & 97.5 & 87.6 & 83.9 & 80.7 & 87.4  \\
 & - & - & (ii) & 97.4 & 91.7 & 84.0 & 84.9 & 89.5  \\
 & - & - & (iii) & 97.6 & 91.2 & 80.9 & 78.4 & 87.0 \\
\midrule 
\multirow{4}{*}{MAML \cite{finn2017model}} & - & - & (i) & 95.6 & 90.1 & 87.0 & 84.5 & 89.3 \\
 & \checkmark & - & (i) & 95.7 & 86.8 & 86.0 & 84.4 & 88.2 \\
 & - & \checkmark & (i) & 93.7 & 85.6 & 82.5 & 78.2 & 85.0 \\
 & \checkmark & \checkmark & (i) & 96.4 & 92.2 & 84.1 & 85.1 & 89.5 \\
\midrule 
\multirow{7}{*}{Reptile \cite{nichol2018first}} & - & - & (i) & 97.2 & 91.3 & 83.7 & 81.1 & 88.3 \\
 & \checkmark & - & (i) & 96.5 & 90.6 & 86.1 & 80.7 & 88.5 \\
 & - & \checkmark & (i) & \textbf{98.1} & 89.5 & 85.8 & 82.9 & 89.1 \\
 & \checkmark & \checkmark & (i) & 97.2 & 93.2 & \textbf{87.9} & \textbf{86.2} & \textbf{91.1} \\
 & \checkmark & \checkmark & (iv) & 96.3 & 92.0 & 85.6 & 84.8 & 89.7 \\
 & \checkmark & \checkmark & (v) & 97.3 & 93.0 & 87.1 & 85.1 & 90.6 \\
 & \checkmark & \checkmark & (vi) & 97.6 & \textbf{93.3} & 87.6 & 86.0 & \textbf{91.1} \\
\midrule
\multicolumn{9}{c}{OSCR} \\
\midrule
\multirow{3}{*}{ERM \cite{naumovich1998statistical}} & - & - & (i) & 97.1 & 83.8 & 71.1 & 72.0 & 81.0  \\
 & - & - & (ii) & 96.7 & 84.3 & 72.3 & 75.3 & 82.2  \\
 & - & - & (iii) & \textbf{97.3} & 84.2 & 71.2 & 69.6 & 80.6 \\
\midrule 
\multirow{4}{*}{MAML \cite{finn2017model}} & - & - & (i) & 94.9 & 81.4 & 72.7 & 77.8 & 81.7 \\
 & \checkmark & - & (i) & 94.0 & 76.9 & 75.7 & 74.7 & 80.3 \\
 & - & \checkmark & (i) & 89.3 & 75.1 & 68.3 & 69.4 & 75.5 \\
 & \checkmark & \checkmark & (i) & 93.4 & 82.9 & 76.2 & 77.1 & 82.4 \\
\midrule 
\multirow{7}{*}{Reptile \cite{nichol2018first}} & - & - & (i) & 96.9 & 84.2 & 72.4 & 73.7 & 81.8 \\
 & \checkmark & - & (i) & 96.1 & 81.2 & 76.2 & 73.1 & 81.7 \\
 & - & \checkmark & (i) & 94.8 & 84.1 & 75.8 & 76.5 & 82.8 \\
 & \checkmark & \checkmark & (i) & 96.9 & 85.3 & 77.2 & 79.2 & 84.7 \\
 & \checkmark & \checkmark & (iv) & 95.8 & 85.0 & 76.1 & 77.9 & 83.7 \\
 & \checkmark & \checkmark & (v) & 97.1 & 85.4 & 76.2 & 78.9 & 84.4 \\
 & \checkmark & \checkmark & (vi) & 97.0 & \textbf{85.8} & \textbf{78.4} & \textbf{79.6} & \textbf{85.2} \\
\bottomrule
\end{tabular}
}
\begin{tablenotes}
    \item[1] Whether the tasks are divided based on different domains or not.
    \item[2] Whether the tasks are divided based on different classes or not.
    \item[3] \change{(i) Random split. (ii) Equal sampling quantity between domains and classes. (iii) Equal output activation regulation. (iv) Task sampling from \cite{luna2020information}. (v) Adaptive task sampling from \cite{liu2020adaptive}, where confusable class pairs are selected but not assigned to different steps. (vi) Our adaptive task sampling strategy, where confusable class pairs are selected and assigned to different steps. }
\end{tablenotes}
\end{threeparttable}
\vspace{-0.1in}
\label{tab: strategy-cmp}
\end{table}

\subsection{Analysis \& Discussion}
\label{subsec:experiment-analysis}

\begin{figure}[tp]
    \centering  
    \vspace{0.1in}
    \changefigure{\includegraphics[width=0.85\linewidth]{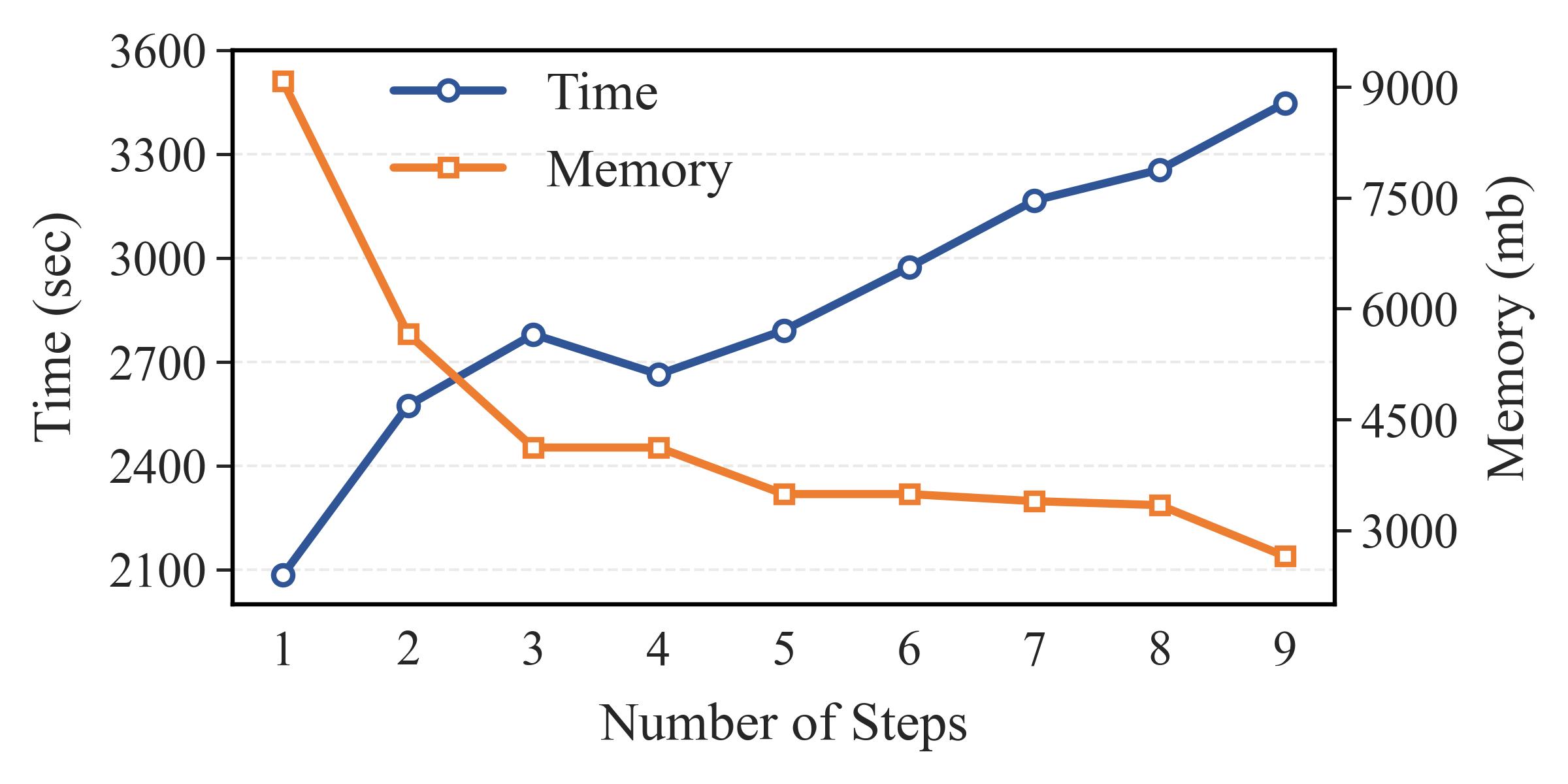}}
    \vspace{-0.1in}
    \caption{Training time (\emph{sec}) and memory cost (\emph{mb}) with respect to the number of steps during 5000 iterations.}
    \vspace{-0.05in}
    \label{fig:time-memory}
\end{figure}

\noindent \change{\textbf{Time and Memory Costs.}  
The primary cost across different steps is a trade-off between time and memory. Model training mainly involves four operations: \emph{forward}, \emph{backward}, \emph{step}, and \emph{zero\_grad}. Most computation occurs in \emph{forward} and \emph{backward}, whereas \emph{step} and \emph{zero\_grad} are lightweight. In MEDIC++, since the total data processed in the \emph{forward} and \emph{backward} passes is equivalent to that of ERM, the overall computational cost remains comparable.
\cref{fig:time-memory} illustrates how the training time and memory utilization change with the number of steps, measured over 5000 iterations on a single Nvidia RTX 2080Ti GPU with a per-task batch size of 8.
In practice, training time increases as the number of steps grows, because smaller batch size can reduce GPU utilization and incur higher kernel launch overhead. However, since each batch's computational graph is released after \emph{backward} pass, peak memory usage is reduced, leading to improved memory efficiency.}

\vspace{0.3em}

\begin{figure*}[t]
    \centering
    \includegraphics[width=0.9\linewidth]{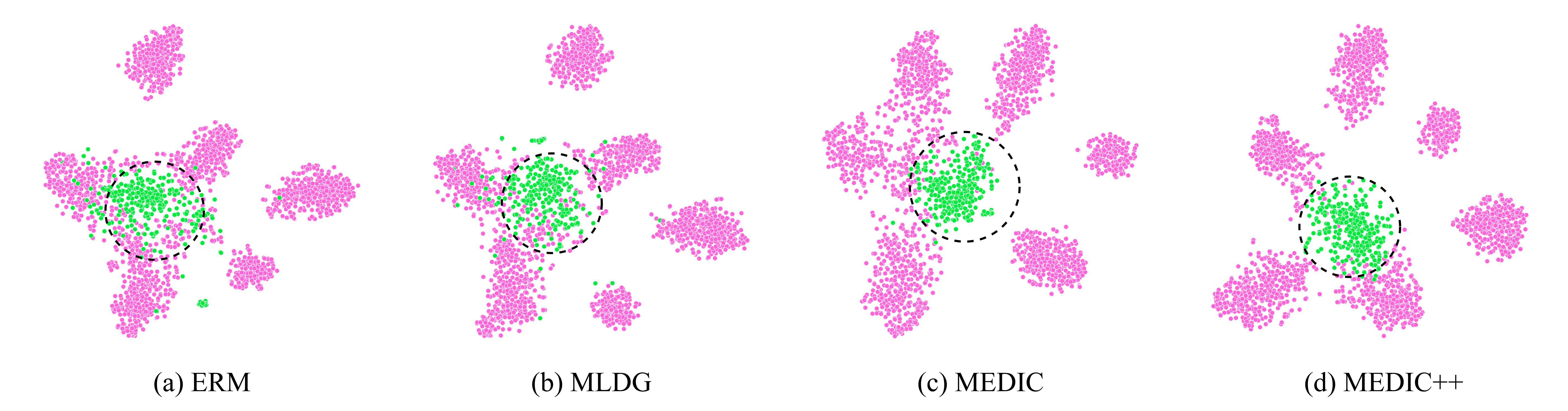}
    \vspace{-0.1in}
    \caption{T-SNE results of feature representations in the target domain, where pink and green corresponds to known and unknown classes respectively.}
    \vspace{-0.05in}
    \label{fig:tsne}
\end{figure*}

\noindent \change{\textbf{Verify unbiased decision boundaries.}
We adopt confidence score to reflect the model’s decision tendency. As illustrated in \cref{tab:conf-cmp}, conf$_{\rm p}$ and conf$_{\rm n}$ denote the average activation of the multi-binary classifier on the positive and negative output channels.
It can be observed that
ERM \cite{naumovich1998statistical}, MLDG \cite{li2018learning}, and Fish \cite{shi2021gradient} exhibit reduced confidence for positive samples than for negative samples, which explains the pattern in left panel of \cref{fig:bias}. In contrast, MEDIC and MEDIC++ produce comparable activation magnitudes for both classes, corresponding to the right panel. These results suggest that inter-class gradient matching effectively promotes unbiased predictions.
We also provide t-SNE feature results in \cref{fig:tsne}.
It can be seen that the unknown classes are generally clustered around the centralized region.
For both MEDIC and MEDIC++, the overlap between known and unknown classes seems smaller, with more space allocated for potential unknown classes.
Compared to MEDIC, MEDIC++ has clearer boundaries across known classes, which helps to explain its superior close set performance.}

\vspace{0.3em}

\noindent \change{\textbf{Discuss unbiased decision boundaries.}
We then discuss why a classifier that exhibits balanced output may improve open set behavior, as it induces maximal uncertainty in regions between known classes under mild linear assumptions.
To simplify the problem, consider binary classification with 
two classes $\mathcal{C}_0$ and $\mathcal{C}_1$ 
and model $f: \mathbb{R}^d \to [0,1]$, which assigns label 0 to $\mathcal{C}_0$ and label 1 to $\mathcal{C}_1$.
Maximum predictive uncertainty occurs when $f(x)=0.5$.
We define balanced output as the sum of average predictions for both classes equals 1:
\begin{equation}
\frac{1}{|\mathcal{C}_0|}
\sum_{x^{c_0}\in \mathcal{C}_0} f(x^{c_0})
+
\frac{1}{|\mathcal{C}_1|}
\sum_{x^{c_1}\in \mathcal{C}_1} f(x^{c_1})
= 1.
\label{eq:1}
\end{equation}
Consider two correctly classified samples 
$x^{c_0}_i \in \mathcal{C}_0$ and $x^{c_1}_j \in \mathcal{C}_1$.
We examine the midpoint as:
\begin{equation}
x_{ij} = \frac{x^{c_0}_i + x^{c_1}_j}{2}.
\label{eq:2}
\end{equation}
Such points lie between known classes and are thus plausible 
candidates for unknown class regions.
We define the average midpoint prediction as:
\begin{equation}
\lambda(\mathcal{C}_0, \mathcal{C}_1, f)
=
\frac{1}{|\mathcal{C}_0||\mathcal{C}_1|}
\sum_{x^{c_0} \in \mathcal{C}_0} \sum_{x^{c_1} \in \mathcal{C}_1}
f\!\left(
\frac{x^{c_0}+x^{c_1}}{2}
\right).
\label{eq:3}
\end{equation}
Smaller $|\lambda - 0.5|$ indicates higher average uncertainty in inter-class regions. Assume that $f$ is locally linear such that, along the segment connecting two samples, it admits that:
\begin{equation}
f\!\left(
\frac{x^{c_0}+x^{c_1}}{2}
\right)
\approx
\frac{1}{2}
\left(
f(x^{c_0}) + f(x^{c_1})
\right).
\label{eq:4}
\end{equation}
Under this approximation,
\begin{equation}
\begin{split}
\lambda
& \approx \frac{1}{2|\mathcal{C}_0||\mathcal{C}_1|}\sum_{x^{c_0} \in \mathcal{C}_0} \sum_{x^{c_1} \in \mathcal{C}_1}\left( f(x^{c_0}) + f(x^{c_1}) \right) \\
&= \frac{1}{2}
\left(
\frac{1}{|\mathcal{C}_0|}
\sum_{x^{c_0} \in \mathcal{C}_0} f(x^{c_0})
+
\frac{1}{|\mathcal{C}_1|}
\sum_{x^{c_1} \in \mathcal{C}_1} f(x^{c_1})
\right).
\end{split}
\label{eq:5}
\end{equation}
Therefore, $|\lambda - 0.5|$ is minimized when \cref{eq:1} holds.
This suggests that balanced output across known classes serves as a sufficient condition for inducing maximal uncertainty in inter-class regions under local linear assumption. 
We emphasize that this argument provides intuition 
rather than a strict guarantee, as the output of deep networks is highly likely nonlinear.}

\vspace{0.3em}

\noindent \textbf{Sharing parameters between classifiers.}
Each known class is linked to a unique one-vs-all classifier and a single output channel in the close set classifier. The function of this channel is similar to that of the positive output channel in the corresponding binary classifier, which is activated by samples from the same class. This raises the question of whether parameter sharing between these channels is feasible. As demonstrated in \cref{tab: param-share}, sharing parameters yields performance comparable to the original architecture, while reducing the total number of output channels from $3|\mathcal{C}|$ to $2|\mathcal{C}|$.

\vspace{0.3em}

\noindent \change{\textbf{Partial domain generalization.}
We consider the presence of domain-specific classes, where splitting domain and class over the entire set is not recommended, as this may create invalid pairs.
In \cref{alg:medic++}, we thus first split by domain in the third line, and then partition within each domain-specific class set in the fifth line.
As illustrated in \cref{tab: pacs6-partial4}, each source domain consists of four known classes, with any two domains sharing two classes and differing in the other two (\ie, $\{1,2,3,4\}$ for domain $1$, $\{1,2,5,6\}$ for domain $2$, $\{3,4,5,6\}$ for domain $3$). MEDIC++ continues to achieve strong performance, indicating the potential of this finer-grained task-level balance approach for other problem settings.}

\vspace{0.3em}

\noindent \change{\textbf{Single domain generalization.}
In this scenario, the model is trained on a single domain and evaluated on all other domains. Our method then reduces to pure class-wise gradient matching, while MLDG and Fish degenerate to standard ERM. We thus compare with CrossMatch \cite{zhu2021crossmatch}, which also uses multi-binary classifiers for open set single domain generalization. As shown in \cref{tab: pacs6-single}, MEDIC++ still performs well in this setting. Since no inter-domain tasks is involved, the gain comes directly from class-wise gradient matching.}

\begin{table}[t]
\caption{Confidence scores (\%) on PACS / sketch.}
\vspace{-0.05in}
\centering
\begin{threeparttable}
\changetable
\begin{tabular}{lccccc}
\toprule
\textbf{Method} & ERM & MLDG & Fish & MEDIC & MEDIC++ \\

\midrule
conf$_{\rm p}$  & 75.59 & 77.30 & 75.76 & 82.23 & 85.08 \\ 
conf$_{\rm n}$  & 84.29 & 84.04 & 85.12 & 82.12 & 85.15 \\ 

\bottomrule
\end{tabular}
\end{threeparttable}
\vspace{-0.1in}
\label{tab:conf-cmp}
\end{table}

\section{Conclusion}
\label{sec:conclusion}
In this paper, we introduce the problem setting of open set domain generalization, which aims to tackle both challenges of domain shift and category shift in the unseen target domain.
We propose a simple yet powerful meta-learning-based framework, which incorporates domain-wise and class-wise gradient matching simultaneously, accompanied by a multi-binary classifier to learn a balanced decision boundary for each known class.
We conduct experiments on multiple benchmarks to demonstrate the superior performance of our approach in both close set and open set scenarios.

\bibliographystyle{IEEEtran}
\bibliography{_references}{}

\end{document}